\documentclass[conference]{IEEEtran}
\IEEEoverridecommandlockouts
\usepackage{cite}
\usepackage{amsmath,amssymb,amsfonts}
\usepackage{algorithmic}
\usepackage{booktabs}
\usepackage{graphicx}
\usepackage{textcomp}
\usepackage{xcolor}
\usepackage{xspace}
\usepackage{amsthm}
\usepackage{indentfirst}
\usepackage{amsmath} 
 \usepackage{mathrsfs}
\usepackage{graphicx}
\usepackage{makecell}
\usepackage{subfigure}
\usepackage{stmaryrd}
\usepackage[ruled, vlined, linesnumbered]{algorithm2e}
\usepackage{multirow}
\usepackage{xcolor}
\usepackage{setspace}
\usepackage{flushend}
\usepackage{url}
\usepackage{float}
\usepackage{balance}

\newtheorem{definition}{Definition}

\newcommand{\eg}{\emph{e.g.},\xspace}
\newcommand{\ie}{\emph{i.e.},\xspace}

\newcommand\figref[1]{Fig.~\ref{#1}}

\newcommand\tabref[1]{Table~\ref{#1}}

\newcommand\secref[1]{Sec.~\ref{#1}}
\newcommand\equref[1]{Eq.(\ref{#1})}

\newcommand\algref[1]{Algorithm~\ref{#1}}
\newcommand{\fakeparagraph}[1]{\vspace{1mm}\noindent\textbf{#1.}}

\ifodd 1

\else

\fi

\def\BibTeX{{\rm B\kern-.05em{\sc i\kern-.025em b}\kern-.08em
    T\kern-.1667em\lower.7ex\hbox{E}\kern-.125emX}}
\begin{document}

%\title{Dynamic and Large-Scale Picking Delivery at Indoor Space}
\title{Adaptive Task Planning for Large-Scale \\ Robotized Warehouses}

\author{
\IEEEauthorblockN{ 
  Dingyuan Shi\IEEEauthorrefmark{1}, Yongxin Tong\IEEEauthorrefmark{1},  Zimu Zhou\IEEEauthorrefmark{2}, Ke Xu\IEEEauthorrefmark{1}, Wenzhe Tan\IEEEauthorrefmark{3}, Hongbo Li\IEEEauthorrefmark{3}}
\IEEEauthorblockA{ 
    \IEEEauthorrefmark{1} \textit{SKLSDE Lab, BDBC and IRI, Beihang University, Beijing, China}\\
    \IEEEauthorrefmark{2} \textit{Singapore Management University} \\
    \IEEEauthorrefmark{3} \textit{Geekplus} \\
	\IEEEauthorrefmark{1}\{chnsdy, yxtong, kexu\}@buaa.edu.cn}  
	\IEEEauthorrefmark{2}zimuzhou@smu.edu.sg
	\IEEEauthorrefmark{3}\{wenzhe.tan, jason.li\}@geekplus.com
}

\SetKwInOut{Input}{Input}
\SetKwInOut{Output}{Output}
\newcommand{\sysname}{EATP\xspace}

\maketitle

\begin{abstract}
Robotized warehouses are deployed to automatically distribute millions of items brought by the massive logistic orders from e-commerce.
A key to automated item distribution is to plan paths for robots, also known as task planning, where each task is to deliver racks with items to pickers for processing and then return the rack back.
Prior solutions are unfit for large-scale robotized warehouses due to the inflexibility to time-varying item arrivals and the low efficiency for high throughput.
In this paper, we propose a new task planning problem called TPRW, which aims to minimize the end-to-end makespan that incorporates the entire item distribution pipeline, known as a fulfilment cycle.
Direct extensions from state-of-the-art path finding methods are ineffective to solve the TPRW problem because they fail to adapt to the bottleneck variations of fulfillment cycles.
In response, we propose Efficient Adaptive Task Planning, a framework for large-scale robotized warehouses with time-varying item arrivals. 
It adaptively selects racks to fulfill at each timestamp via reinforcement learning, accounting for the time-varying bottleneck of the fulfillment cycles.
Then it finds paths for robots to transport the selected racks.
The framework adopts a series of efficient optimizations on both time and memory to handle large-scale item throughput.
Evaluations on both synthesized and real data show an improvement of $37.1\%$ in effectiveness and $75.5\%$ in efficiency over the state-of-the-arts.
\end{abstract}

% \begin{IEEEkeywords}
% k1, k2, k3
% \end{IEEEkeywords}

\section{Introduction}
\label{sec:intro}

The boom of e-commerce has stimulated enormous logistic demands.
Over 2 billion logistic orders (worth over 115 billion dollars) were created during the online shopping carnival of 2020 in China\footnote{https://www.cnbc.com/2020/11/12/singles-day-2020-alibaba-and-jd-rack-up-record-115-billion-of-sales.html}.
Such huge amounts of orders often emerge dynamically over time. 
For example, there can be a sharp surge within a short time when the carnival begins at midnight.
The massive, time-varying arrival of orders in unit time \ie throughput, urges highly efficient and effective operations of the warehouses that store and distribute the corresponding items to buyers \cite{IR16Robert}.

\begin{figure}[t]
    \centering
    \subfigure[]{
		\includegraphics[width=0.45\linewidth]{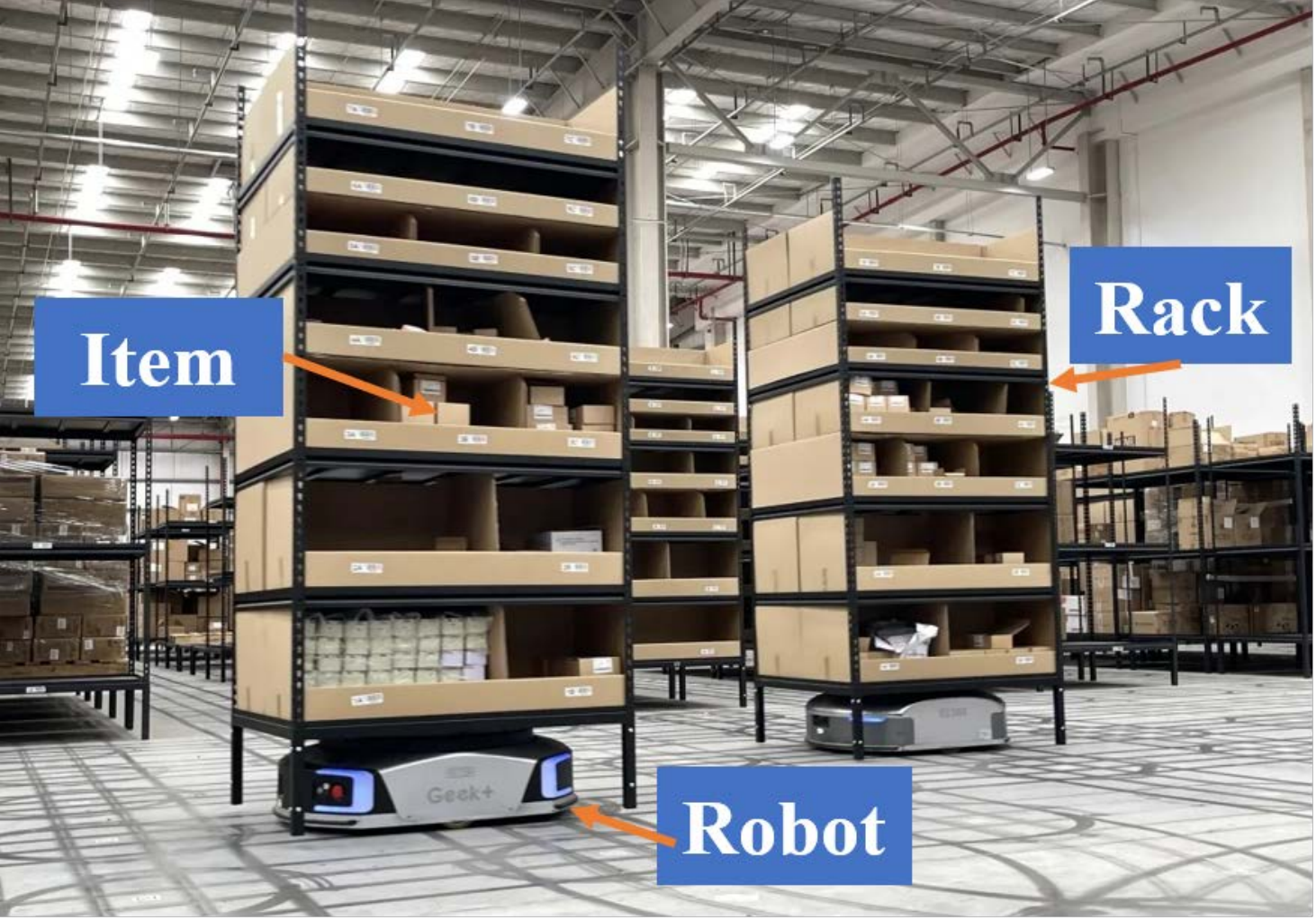}
		\label{fig:robot}
	}
	\subfigure[]{
		\includegraphics[width=0.45\linewidth]{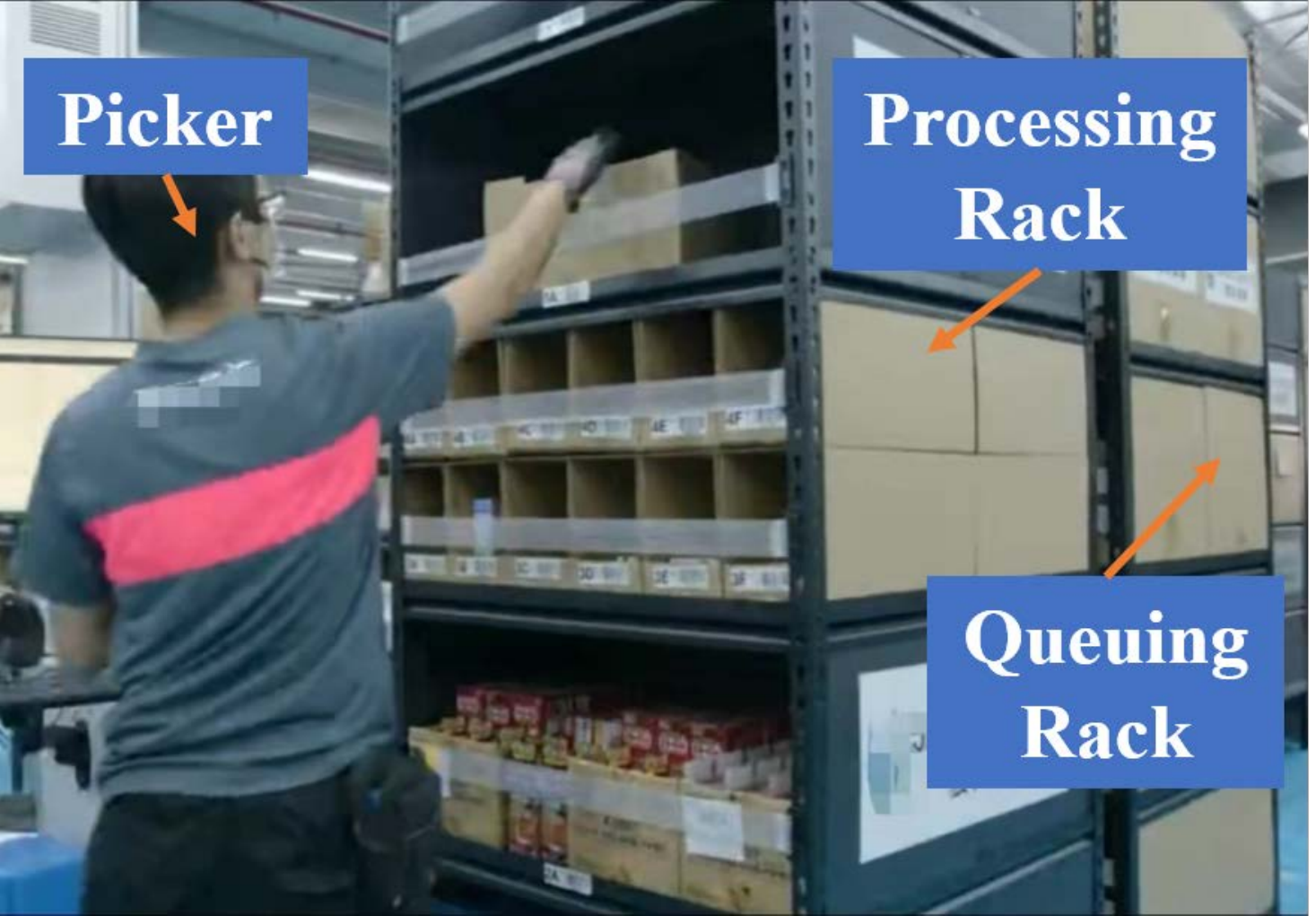}
		\label{fig:picker}
	}
	\caption{A snapshot of a robotized warehouse showing entities in (a) the storage area and (b) the processing area.}
	\label{fig:entity}
\end{figure}

\begin{figure}[t]
    \centering
    \includegraphics[width=0.8\linewidth]{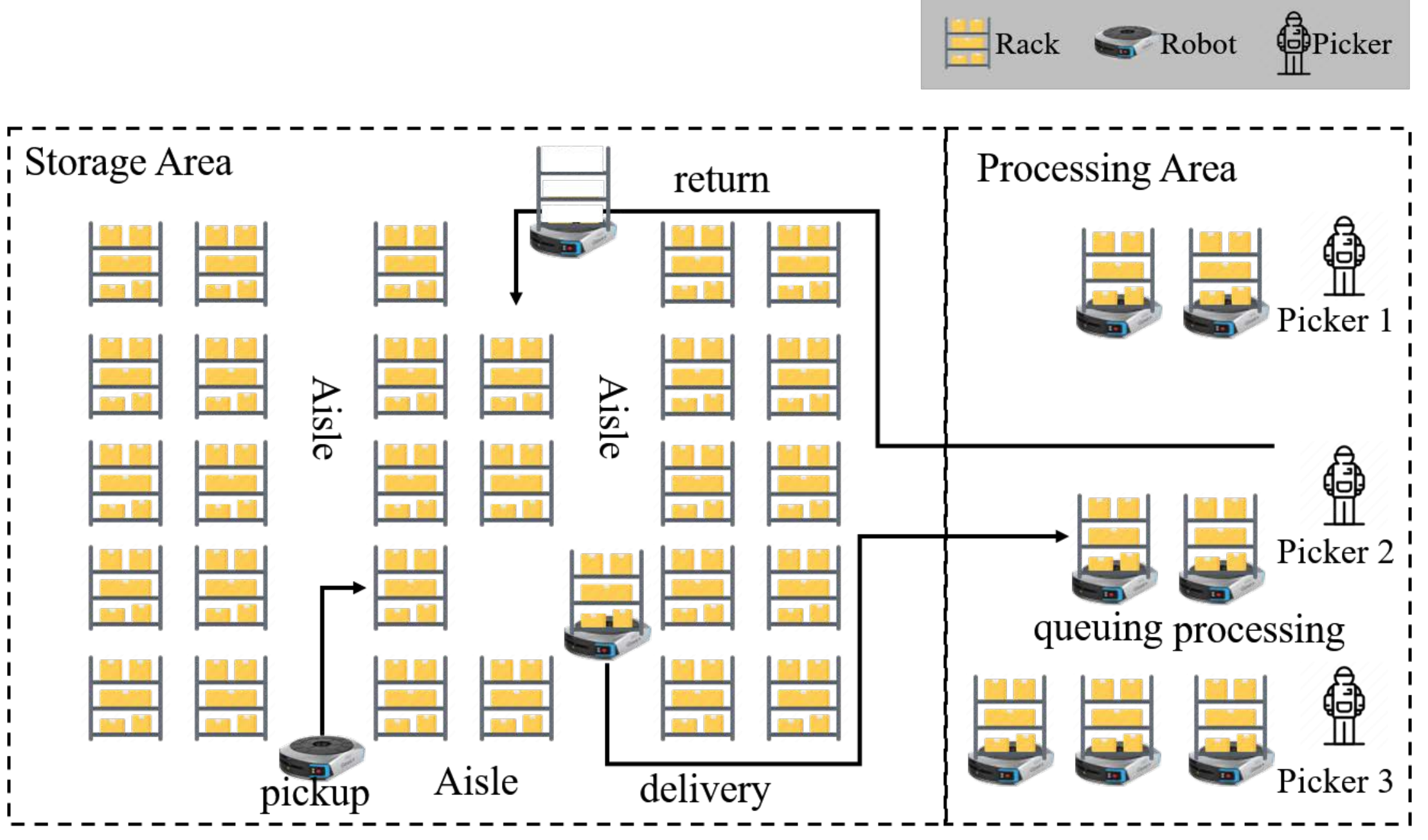}
    \caption{The 2D layout of a rack-to-picker warehouse, where racks are shipped back and forth between storage area and picking area. 
    Pickers located at picking area processing items (tasks) on the racks.
    A complete fulfilling cycle contains five steps: pickup, delivery, queuing, processing and return.}
    \label{fig:warehouse}
\end{figure}

Robotized warehouses are expected to improve the effectiveness and efficiency of warehouse operations by automating the fulfillment cycle of item distributions \cite{AI08Wurman}.
In these warehouses, multi-robot systems are installed for item distribution. 
Of our particular interest is the rack-to-picker mode, a popular robotized warehouse operational mode where robots pick up and deliver racks containing items from the storage area to pickers in the human picker area for processing (see \figref{fig:entity}).
In this mode, a fulfill cycle for item distribution consists of five steps: rack pickup, delivery, queuing, processing, and return (see \figref{fig:warehouse}). 
From the algorithmic perspective, a central problem is to plan tasks (\ie items) for these robots, \ie determine racks (containing tasks) to fulfill, and plan paths for robots to complete fulfill cycles at each timestamp.

Such task planning problems have been extensively studied in the context of multi-agent path finding \cite{AAAI10Standley, AI13Sharon, AI15Sharon, IJCAI15Boyarski, AAMAS17Ma, AAAI19vancara, AAAI21Li}. 
These multi-agent path finding algorithms search conflict-free paths for multiple agents (\ie no robot will collision with each other), typically with the objective to find the shortest paths \cite{AAAI10Standley, AI13Sharon}, or to minimize the makespan \cite{AI15Sharon, IJCAI15Boyarski, AAAI19vancara, AAMAS17Ma, AAAI21Li}, \ie the total delay to fulfill all items \cite{SOCS19Stern}.
Most research efforts perform \textit{offline} task planning assuming the arrival of items is known a prior \cite{AAAI10Standley, AI13Sharon, AI15Sharon, IJCAI15Boyarski}.
A few \cite{AAMAS17Ma, AAAI19vancara, AAAI21Li} investigate the more realistic \textit{online} task planning problem, where tasks come continually as times goes on.
We also focus on online task planning that minimizes the makespan.
Yet we argue that prior studies \cite{AAMAS17Ma, AAAI19vancara, AAAI21Li} are unfit for online task planning in large-scale robotized warehouses due to the following limitations.
\begin{itemize}
    \item 
    \textit{Limitation 1: inflexible planning to time-varying item arrival.} 
    As mentioned, a fulfillment cycle contains multiple stages.
    Previous studies \cite{AAMAS17Ma, AAAI19vancara, AAAI21Li} assume a fixed makespan bottleneck, \eg delivery, which is reasonable with low, constant throughput. 
    This assumption breaks with high, varying throughput, where the makespan bottleneck may turn into queuing or processing.
    It is ineffective to apply a time-invariant planning strategy to cope with the dynamic makespan bottleneck.
    \item 
    \textit{Limitation 2: inefficient planning for massive robots and items.}
    Many path finding algorithms \cite{AAMAS17Ma, AAAI19vancara, AAAI21Li} adopt A* search \cite{TSSC68Peter} from source to destination \textit{completely}, often resulting in a time complexity of $O(I(HW)^2)$, where $I, H, W$ is the number of items, height and width of warehouse.
    However, modern warehouses for e-commerce are confronted with million-scale processing workload and thousands of robots\footnote{https://www.dhl.com/nl-en/home/press/press-archive/2019/dhl-opens-largest-and-greenest-e-commerce-sorting-center-for-the-dutch-market.html}.
    With over $10^6$ items flushing in, the total time complexity will be up to $10^{14}$, which is unacceptable for execution in practice.
    Therefore, more efficient planning algorithms are compulsory.
\end{itemize}

%\dy{
In response, we take a holistic problem formulation.
First, we define an end-to-end makespan incorporating the entire fulfillment cycle.
Such a formulation captures the bottleneck changes in fulfillment cycles due to the time-varying item arrivals.
%Direct extensions from current state-of-the-art methods are not capable of solving this problem due to the inability to bottleneck variations of fulfillment cycles.
%}
Then we propose Efficient Adaptive Task Planning (\sysname), an effective and efficient task planning framework for large-scale robotized warehouses with time-varying item arrivals.
Instead of start fulfilling once items emerge on racks, \sysname adopts reinforcement learning to adaptively select racks to fulfill at each timestamp according to the current throughput, where the makespan may be dominated by the delay of rack transport (pickup, delivery, and return), processing, or queuing.
It also incorporates a series of efficient designs such as flip requesting side, conflict detection table and cache aiding for both time and memory consumption reduction in terms of rack selection and path finding.
Evaluations on both synthesized and real data show an improvement of $37.1\%$ in effectiveness and $75.5\%$ in efficiency over the state-of-the-art online multi-agent path finding scheme \cite{AAMAS17Ma}.

Our contributions are summarized as follows.
\begin{itemize}
    \item 
    We formulate the Task Planning in Robotized Warehouse (TPRW) problem, which aims to minimize the end-to-end makespan and highlights the challenges with massive, time-varying item arrivals.
    \item 
    We design Efficient Adaptive Task Planning (\sysname), an effective and efficient task planning framework to solve the TPRW problem.
    It offers adaptability to item arrivals with reinforcement learning based rack selection, and adopts a series of optimizations for fast and scalable multi-agent path finding.
    \item 
    We conduct experiments on both synthesized and real datasets. 
    The results demonstrate notable gains over the state-of-the-art \cite{AAMAS17Ma} in both effectiveness and efficiency.
\end{itemize}

The rest of this paper is organized as follows.
We first formulate our problem in \secref{sec:problem} and propose a naive task planning algorithm in \secref{sec:baseline}.
We then provide an overview of our efficient adaptive task planning method in \secref{sec:overview}, and elaborate on its learning based rack selection and robot path finding in \secref{sec:atp}.
The efficient design are detailed in \secref{sec:efficient}.
We present the evaluations in \secref{sec:evaluation}, review related work in \secref{sec:related}, and finally conclude in \secref{sec:conclusion}.

\section{Problem Statement}
\label{sec:problem}

In this section, we define the \underline{T}ask \underline{P}lanning in \underline{R}obotized \underline{W}arehouses (TPRW) problem.
We formulate the problem in the context of the rack-to-picker mode, a prevailing operational mode in robotized warehouses \cite{EJOR17Boysen}.
In this operation mode, \textit{robots} ship \textit{racks} containing multiple items from the storage area to \textit{pickers} in the processing area.
% The task scheduling problem consists of two sub-problems: 
% \textit{(i)} select racks to deliver and assign them to robots and \textit{(ii)} plan paths for robots to pick up, deliver, and return racks.
As in prior studies \cite{AAMAS17Ma, AAAI19vancara, AAAI21Li}, we partition the warehouse into grids whose side length is the same as a robot's side length (about 1 meter).
The grid partition is reasonable because the layout of a warehouse is often regular.
We build grid index for the warehouse. 
%Note that a warehouse is no more than 500 meters wide so grid index will not take too much space.
Next, we formally define racks, pickers, and robots.

\begin{definition}[Rack]
\label{def:rack}
A rack $r$ is represented as $\langle l_r, \tau_r, p_r \rangle$, which locates at $l_r$ in the storage area and is associated with picker $p_r$.
$\tau_r$ is the set of items' processing time unit consumption on $r$ to be delivered to $p_r$.
\end{definition}
In the rack-to-picker mode, each rack is associated with a fixed picker.
This is because certain pickers and racks may be dedicated to serve items destined to specific cities.
Item processing time usage set $\tau_r$ specifies each item's processing time at the picker.
% which are estimated from historical data and they are input for our problem.
% which varies across items due to their differences in size or weight.
% The item processing time can be estimated from historical data and we take it as input for our problem.
Items arrive and processed in an online manner thus elements in $\tau_r$ emerge and disappear as time goes on.
We use $R$ to denote the set of all racks.

\begin{definition}[Picker]
\label{def:picker}
A picker $p$ is represented as $\langle l_p, q_p, e_p \rangle$, where $l_p$ is its fixed location and $e_p$ is the estimated remaining processing time of the currently picked item.
$q_p$ is the queue of racks waiting to be processed.
\end{definition}
We assume a picker processes the items on the racks in the queue in the ``first-come-first-serve'' manner.
This is reasonable since the picking area is often confined, making it difficult for robots that carry racks to cut in line.
We use $P$ to denote the set of all pickers.

\begin{definition}[Robot]
\label{def:amr}
A robot $a$ is represented as $\langle l_a, s_a\rangle$ with location $l_a$ and state $s_a$.
\end{definition}
Since the robot is mobile, its location and state change over time.
The state $s_a$ can be either busy or idle.
A robot is busy if it is in the stage of rack pickup, delivery, queuing, processing, or return.
We use $A$ to denote the set of all idle robots.

Now we define the objectives of our task planning problem.
In short, we aim to minimize the \textit{makespan}.
\begin{definition}[Makespan]
\label{def:makespan}
The makespan $M$ is the time from the emergence of the first item till the return of the last rack.
\end{definition}
Assuming the first item appears at time $0$, $M$ equals to the time when the last rack is returned:
\begin{equation}
\label{eq:makespan}
    M = \max_{r \in R} f_r
\end{equation}
where $f_r$ denotes the latest time at which rack $r$ is returned.
It can be calculated as follows.
\begin{equation} 
\label{eq:finishRack}
\begin{aligned}
    f_r & = t_k + d(l_a, l_r) + d(l_r, l_{p_r}) \\ & + \max\{d(l_a, l_r) + d(l_r, l_{p_r}) - f_p,  0\} 
    + \sum_{i\in \tau_r}i + d(l_{p_r}, l_r)
\end{aligned}
\end{equation}
where $t_k$ is the last time the rack is selected.
The remaining five terms correspond to the delays for rack pickup, delivery, queuing, processing, and return, respectively.
$d(\cdot, \cdot)$ is the path length between two locations.
Assuming that robots move at unit velocity, $d(\cdot, \cdot)$ equals to the delay.
$f_p$ is the delay of picker $p$ to process the items on all the racks in queue, which can be computed as follows.

\begin{equation}
\label{eq:finishPicker}
    f_p = e_p + \sum_{r \in q_p} \sum_{i \in \tau_r} i
\end{equation}

Makespan is a widely used metric in prior studies \cite{AAAI10Standley, AI13Sharon, AAMAS17Ma, AAAI19vancara},
but we redefine it in an end-to-end manner.
From \equref{eq:makespan} to \equref{eq:finishPicker}, the makespan is determined after deciding planning schemes $U = \{U_t\}$ for the robots at each timestamp $t$, where $U_t$ is the planning scheme generated at timestamp $t$.
Finally, we define the problem below.

\begin{figure}[t]
	\centering
	\subfigure[Single-grid]{
		\includegraphics[width=0.45\linewidth]{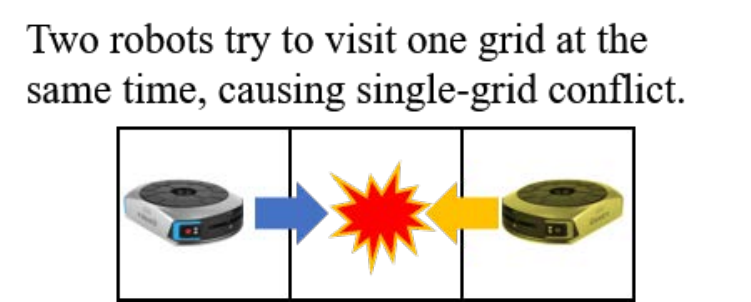}
		\label{fig:expSV}
	}
	\subfigure[Inter-grid]{
		\includegraphics[width=0.45\linewidth]{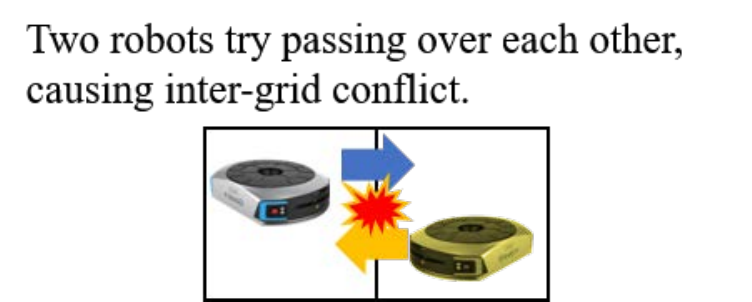}
		\label{fig:expHO}
	}
	\caption{Examples of conflicts in path planning.}
	\label{fig:expConflict}
\end{figure}

\begin{definition}[\underline{T}ask \underline{P}lanning in \underline{R}obotized \underline{W}arehouses (TPRW)]

Given sets of racks $R$, idle robots $A$ and pickers $P$ at every timestamp, the problem is to generate planning schemes $U_t = \{u_a| s_a = \text{idle}\}$ correspondingly at each timestamp in which element $u_a$ is a path for an idle robot $a$, such that
\begin{equation}
\begin{aligned}
    & \min M \\
    & \{U_t| \forall t\} \text{ is sufficient and conflict-free} 
    \nonumber
\end{aligned}
\end{equation}
\end{definition}

The planning schemes $\{U_t| \forall t\}$ are sufficient if all racks are assigned to a robot after the arrival of its items, \ie $\forall i_t \in I_R$, $\exists t' \ge t$ s.t. $\exists a s.t. u_a \in U_{t'}$, where item $i_t$ emerges at time $t$ and $I_R$ is the item set of rack set $R$.

The planning schemes are conflict-free if there is neither \textit{single-grid} conflict nor \textit{inter-grid} conflict among all paths.
The two conflicts are defined below (see \figref{fig:expConflict}).
\begin{itemize}
    \item \textit{Single-grid conflict.} 
    Two paths visit the same location at the same time.
    \item \emph{Inter-grid conflict.} 
    For two paths $u_1$ and $u_2$, there exist $\langle t_i, x_i, y_i \rangle$, $\langle t_{i + 1}, x_{i+1}, y_{i+1} \rangle$ in $u_1$ and $\langle t_j, x_j, y_j \rangle$, $\langle t_{j + 1}, x_{j + 1}, y_{j + 1} \rangle$ in $u_2$ such that $t_i=t_j \land t_{i+1} = t_{j+1} \land x_i=x_{j+1} \land y_i = y_{j+1} \land x_{i+1}=x_j \land y_{i+1} = y_j$.
\end{itemize}

\tabref{tab:notations} summarizes the important notations.

\begin{table}[t]
	\centering
	\caption{Summary of important notations.}
	\label{tab:notations}
	\small
	\begin{tabular}{cl}
	\toprule
		Notation    &   Description	\\	
	\midrule
		$t$ & current timestamp \\
		$R, P, A$	&	set of racks, pickers, idle robots	\\
		$l_r, \tau_r, p_r$ & \makecell[l]{location, item processing time usage set,\\  corresponding  picker of rack $r$} \\
		$l_p, q_p, e_p$ & \makecell[l]{location, rack queue and estimated remaining \\ time of current item of picker $p$} \\
		$l_a/s_a$ & \makecell[l]{location/state of robot $a$} \\
		$M$ & makespan of all tasks \\
		$d(\cdot, \cdot)$ & distance between two locations \\
		$\mathcal{A}$ & designed algorithm \\
		$f_r/f_p$ & finish time of rack $r$/picker $p$ \\
		$U_t, u_a$ & \makecell[l]{path planning scheme at $t$ and a single path \\ for robot $a$} \\ 
		$S_t$ & rack selection scheme at timestamp $t$ \\ 
		$H/W$ & height/width of the warehouse\\ 
		$k, \xi$ & \makecell[l]{number of tasks, processing time of each task \\ 
		in \secref{subsec:analysis}'s example} \\
		$o_i/v_j$ & one task for $p_1$/$p_2$ 	in \secref{subsec:analysis}'s example. \\
		$D/D_j$ & \makecell[l]{summation of pickup, delivery and return time \\ of all $o_i$/each $v_j$ in \secref{subsec:analysis}'s example} \\
		$M$ & \makecell[l]{time usage of moving between $p_1$'s rack and \\ $p_2$'s first rack in \secref{subsec:analysis}'s example} \\ 
		$s, \alpha, \gamma, c, \beta, \epsilon$ & \makecell[l]{state, action, discount factor, reward, \\ learning rate and policy derivation parameter} \\
		$ap_r/ar_r$ & picker $p_r$/rack $r$'s accumulative processing time\\
		$\delta, K, L$ & \makecell[l]{parameters of bootstrap, robot requesting\\and distance threshold } \\
	\bottomrule
	\end{tabular}
\end{table}

\fakeparagraph{Remarks}
The TPRW problem is a variant of the online multi-agent path finding problem \cite{AAMAS17Ma} with an end-to-end makespan definition that accounts for the entire fulfillment cycle (pickup, delivery, queuing, processing, and return).
The TPRW problem is challenging for large-scale warehouses with highly varied item throughput (\ie amount of item arrivals in unit time) due to the complex composition of the makespan.
\begin{itemize}
    \item 
    With low item throughput, the makespan is dominated by the rack transport delay (pickup, delivery, and return) \cite{IR16Robert}, as is optimized by mainstream online multi-agent path finding literature \cite{AAMAS17Ma, AAAI19vancara, AAAI21Li}.
    % \item 
    % With medium item throughput, the accumulative processing time at pickers may surpass that for rack transport, making it the dominating factor in the makespan.
    \item 
    With high item throughput, large queues may build up at pickers, turning the queuing time the bottleneck.
\end{itemize}
In \secref{subsec:case} we will show that the bottleneck changes under different throughput. 
The state-of-the-art solution \cite{AAMAS17Ma} fails to adapt to such bottleneck changes in the makespan, while other studies \cite{AAAI19vancara, AAAI21Li} assume tasks and robots emerge in a binding manner, which is unfit for our problem. 
Neither do they offer efficient design for warehouses with hundreds of robots processing thousands or even millions of items daily.
Next we will show the extension of state-of-the-art solution to our problem and how it may lead to bad results.

\section{Naive Task Planning}
\label{sec:baseline}
In this section, we adapt the state-of-the-art online multi-agent path finding algorithm \cite{AAMAS17Ma} to solve the TPRW problem (\secref{subsec:greedy}), and analyze why it is ineffective (\secref{subsec:analysis}).

\begin{algorithm}[t]
\label{alg:ntp}
    \caption{Naive Task Planning}
	\Input{$t$: timestamp, $P$: pickers, $A$: robots, $R$: racks}
	\Output{$U_t$: planning scheme at $t$}
	$U_t \leftarrow \emptyset$ \\
	Sort $P$ in ascending order based on finishing time $f_p$ \\
    \For{$p \in P$}{
        $R \leftarrow  \{r | \tau_r \ne \emptyset \land p_{r} = p \}$ \\
        \For{$r \in R$}{
	        find the closet idle robot $a$ from $A$\\
	        $u_a \leftarrow \text{ plan path for robot } $a$ \text{via A* algorithm}$ \\
	        $U_t \leftarrow U_t \cup \{u_a \}$ \\
	    }
    }
	\Return{$U_t$}
\end{algorithm}

\subsection{Extension from State-of-the-Art}
\label{subsec:greedy}

The state-of-the-art online multi-agent path finding algorithm \cite{AAMAS17Ma} plans conflict-free paths for each robot one at a time following certain order.
The order of planning is decided by the distance from robots to its closest rack.
That is, this algorithm greedily plans paths for robots with the least pickup time.

This  algorithm is inapplicable to the TPRW problem since it only accounts for pickup and delivery time.
We extend the algorithm to our problem as follows.
Instead of planning paths for robots with the least pickup time, we plan paths for robots associated with the most slack picker.
Specifically, the most slack picker $p$ has the smallest finish time $f_p$ as in \equref{eq:finishPicker}.
This is because a slack picker indicates a smaller queuing time.
Together with optimization on rack transport time, the algorithm tend to minimize the makespan defined in \equref{eq:makespan}.

\algref{alg:ntp} illustrates the naive path planning algorithm.
It first finds all pickers whose corresponding racks that still have items for processing.
Then, it sorts pickers based on their degree of slack (\ie the current finish time).
For each picker, it finds the corresponding racks that require processing.
Finally, it choose the closet idle robot and finds a path via A* algorithm, a classical yet prevailing algorithm in multi-agent path finding studies \cite{AAAI10Standley, AAMAS17Ma, AAAI21Li, AAAI19vancara} (detailed in \secref{subsec:pathfinding}).

\subsection{Limitations of Naive Task Planning}
\label{subsec:analysis}

We illustrate the limitations of the naive task planning algorithm via the following example.

% The greedy algorithm has the time complexity of $O(|P|\log |P| + |R| \cdot |A|)$.
% Though efficient, the algorithm suffers bad performance.
% In fact, the competitive ratio of greedy algorithm is non-constant.
% We build below case for illustration.

\begin{figure}[t]
	\centering
    \includegraphics[width=0.7\linewidth]{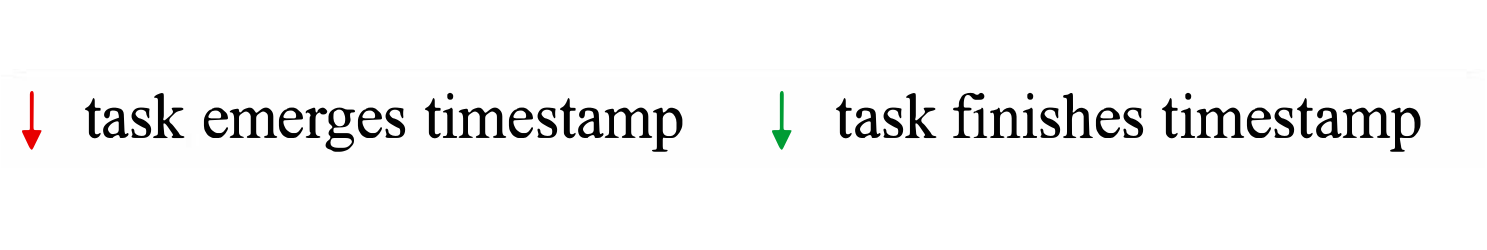} 
	\subfigure[Naive]{
		\includegraphics[width=0.9\linewidth]{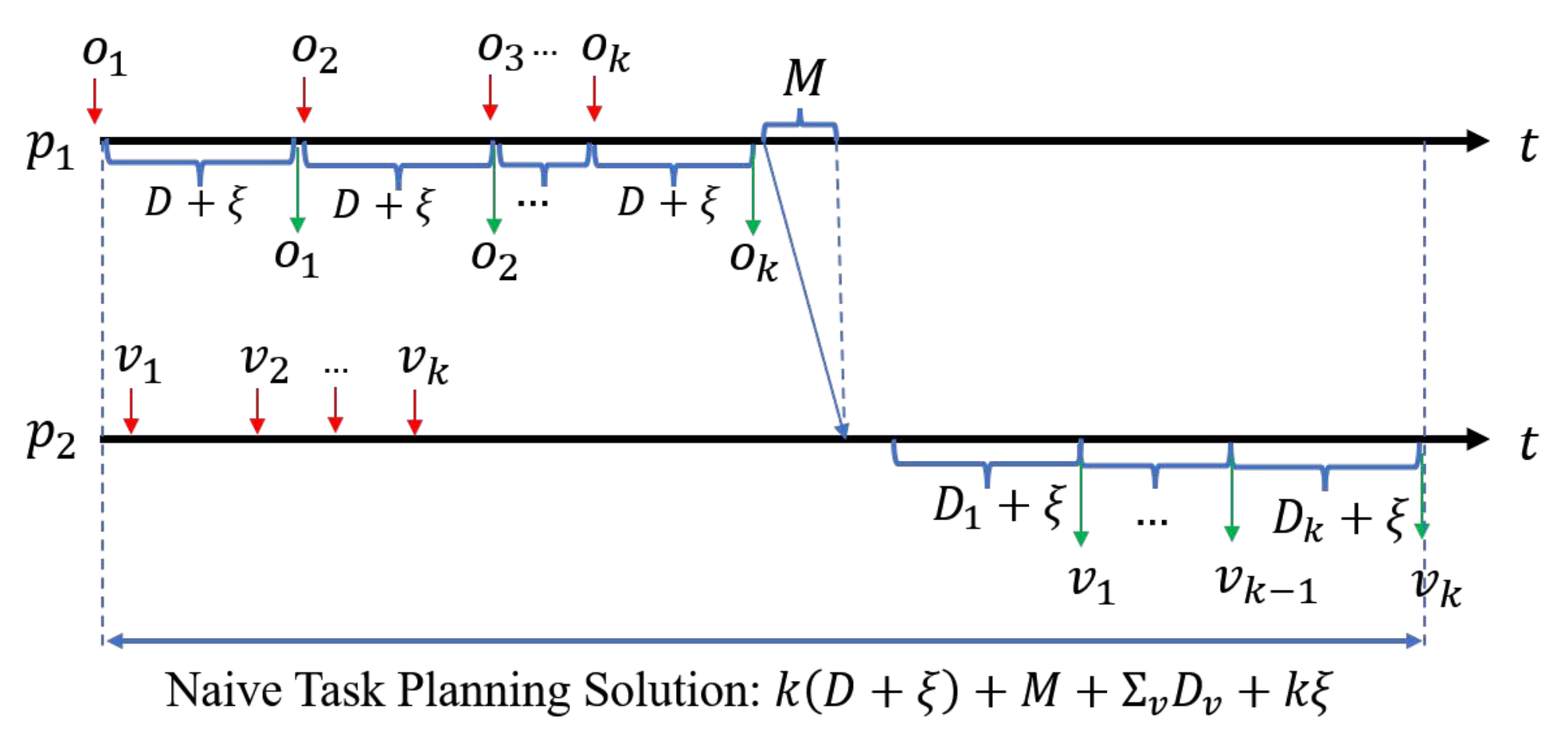}
		\label{subfig:case_worst}
	}
	\subfigure[Optimal]{
		\includegraphics[width=0.9\linewidth]{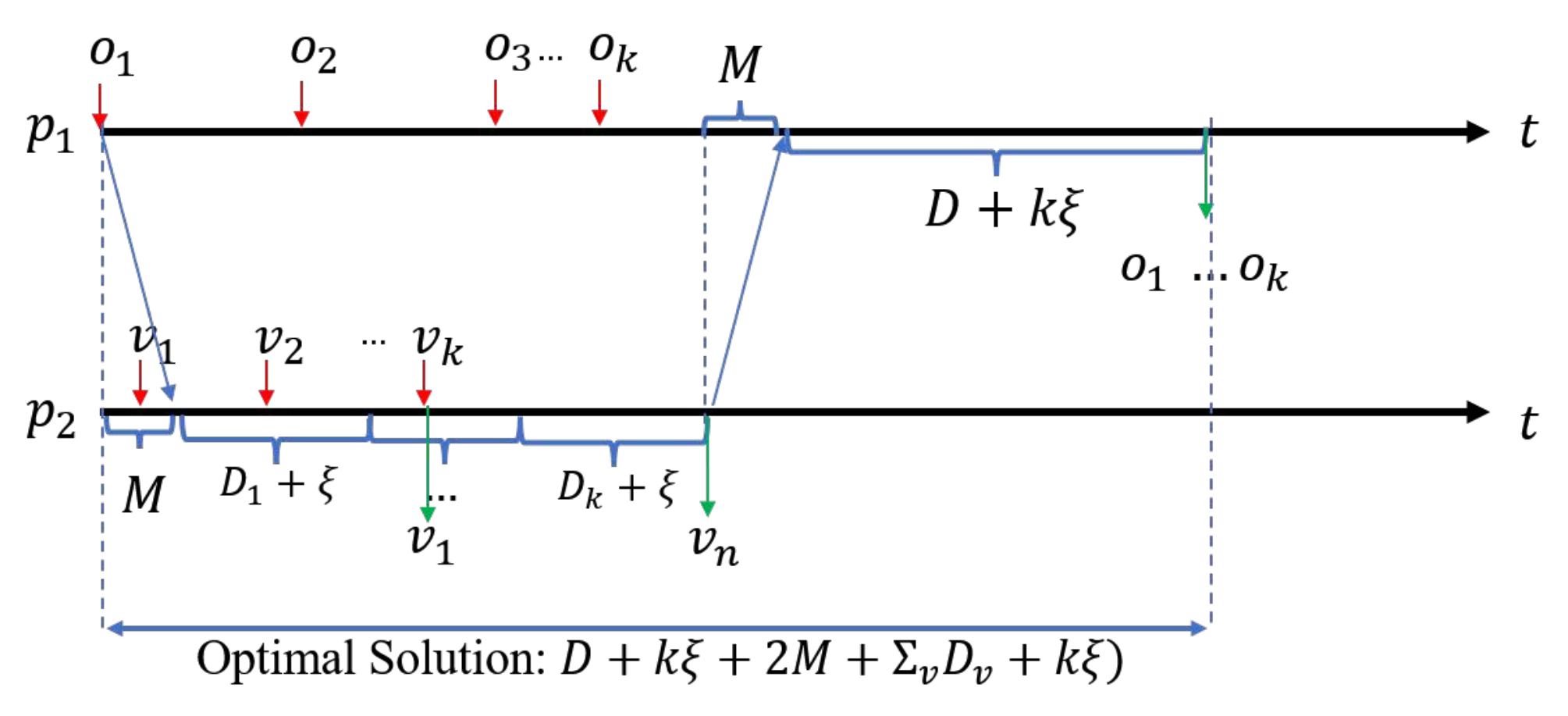}
		\label{subfig:case_opt}
	}
	
	\caption{A bad case for the naive path planning algorithm.}
	\label{fig:case}
\end{figure}

Consider two pickers $p_1, p_2$ and one robot $a$.
For $p_1$, there is only one associated rack $r$.
For $p_2$, $k$ racks on which tasks will emerge.
Robot $a$'s initial location is the same as rack $r$ (right under the rack).
Both $p_1$ and $p_2$ have $k$ items (tasks) to be processed.
For simplicity, we assume all items of $p_1$ and $p_2$ have the same processing time $\xi$.

Picker $p_1$'s all $k$ tasks are $o_1, o_2, ...o_k$ which appear on this rack with the same time interval $D+\xi$, where $D$ is the sum of pickup, delivery and return time from $r$ to $p_1$.
For $p_2$, there are $k$ tasks $v_1, v_2, ...v_k$ and each belongs to a different rack.
Let $D_{j}$ be the sum of pickup, delivery and return time for item $v_j$.
These $k$ items emerge in an online manner and the span between adjacent items $v_i$ and $v_{i+1}$ is shorter than all $D_{i}$.
The emerging time of $v_1$ is later than $o_1$.

In this case, the greedy algorithm will first move rack $r$ to $p_1$ right after the emergence of item $o_1$.
Then when the rack is returned, $o_2$ just shows up and the robot will deliver the rack to $p_1$ again.
The cycle repeats until $o_k$ is finished.
Meanwhile, all the items of $p_2$ have emerged.
The robot then moves from $r$ to the rack of $v_1$, taking $M$ time units.
Then it delivers all racks of $p_2$ one by one, which takes $\sum_{v} D_{v}+k\xi$ time units.
The makespan is $k(D + \xi) + M + \sum_{v} D_{v}+ k\xi$, as shown in \figref{subfig:case_worst}.
The optimal solution, however, will not greedily deliver racks of $p_1$.
It will first deliver all racks of $p_2$.
Meanwhile, all items of $p_1$ will emerge, and it delivers rack $r$ to $p_1$ only once.
The makespan is $D+k\xi + 2M + \sum_{v} D_v+ k \xi$, as shown in \figref{subfig:case_opt}.
Thus, the competitive ratio will be $O(\frac{k(D + \xi) +\sum_{v} D_v+k\xi}{D+k\xi + 2M + \sum_{v} D_v+k\xi})$.
With sufficiently large $D$, the bound is approximately $O(k)$, which is not constant.

The ineffectiveness of the naive path planning algorithm can be explained intuitively as follows.
\begin{itemize}
    \item 
    From picker $p_1$'s perspective, all items emerge on a single rack.
    Hence a smart strategy should be batching the delivery of all items in one time rather than moving racks as soon as one item emerges.
    \item 
    From picker $p_2$'s perspective, all items emerge on different racks.
    So, the bottleneck is the rack transport time.
\end{itemize}
In summary, the difference in throughput \ie the number of items emerged on a rack in unit time, shifts the dominating factor in the makespan, which affects the decision on whether to deliver a rack once an item appears or wait for more items to emerge (experimental results in \secref{subsec:case} also validate the observation).
Naive task planing fails to incorporate such decisions for its greedy strategy. 
This limitation motivates us to develop a solution that adapts its decisions (\ie immediately deliver a rack or wait for more items to arrive) according to the dynamic throughput, as explained next.

% We conclude from above observations that apart from planning short paths for efficient fulfillment, the decision on planning options (\ie which robot and when to execute the fulfilment) also matter.
% Maybe one may infer empirically from the above explanation that maybe items distributed more concentrated (as $p_1$) should be transmit in a low frequency.
% In real situations, however, different items emerge on different racks in an online manner and one picker is responsible for multiple racks.
% Those complexity coupling makes it hard to judge what is the bottleneck that decides the processing efficiency.
% To overcome this, algorithm requires for high adaptibility for planning options, as we will explain next.

\begin{figure}[t]
    \centering
    \includegraphics[width=0.9\linewidth]{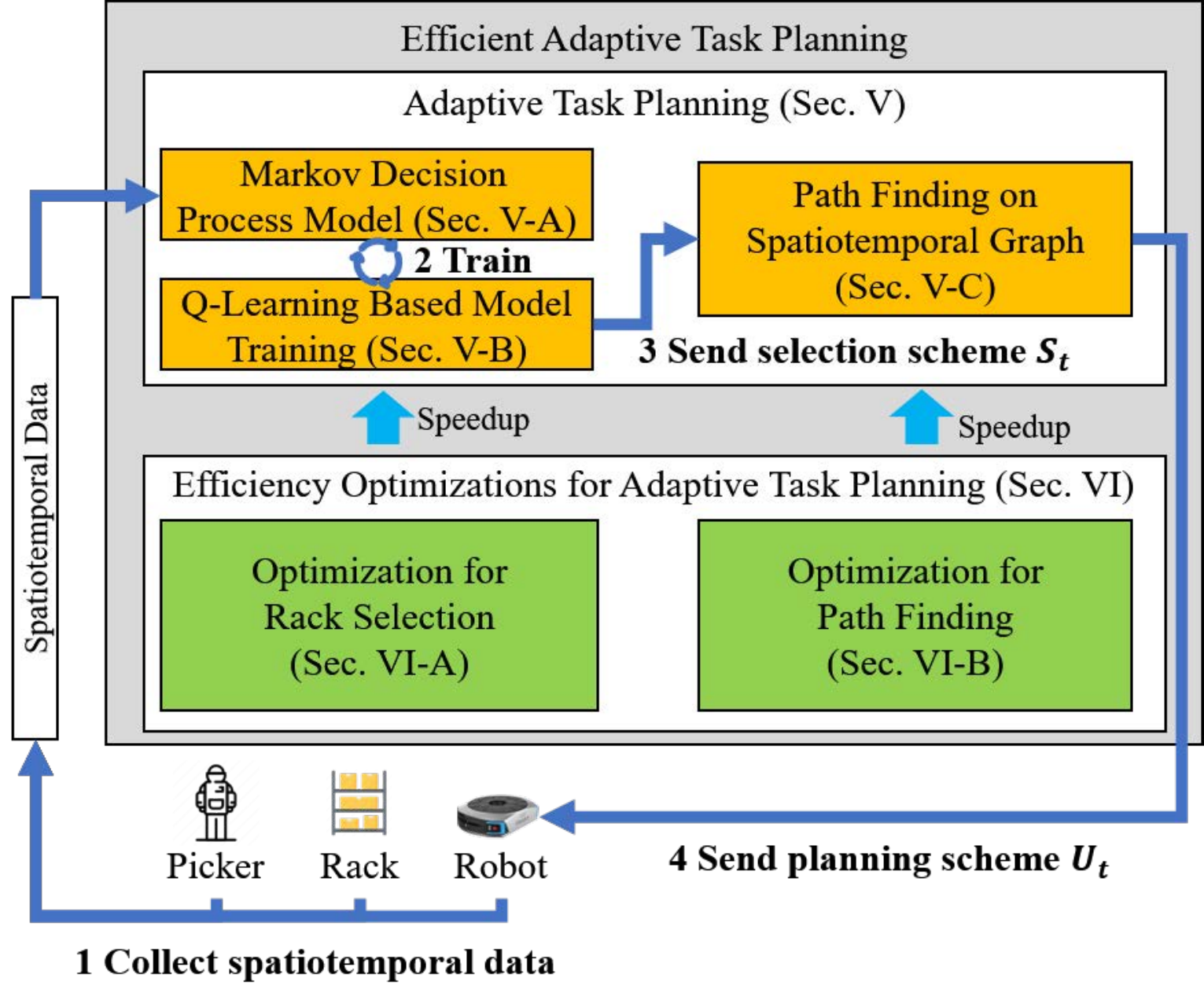}
    \caption{Efficient Adaptive Task Planning (\sysname).}
    \label{fig:system}
\end{figure}

\section{Efficient Adaptive Task Planning Overview}
\label{sec:overview}

This section presents an overview of our \underline{E}fficient \underline{A}daptive \underline{T}ask \underline{P}lanning (\sysname) solution (see \figref{fig:system}).
From the analysis of naive task planning in \secref{subsec:analysis}, its poor performance is due to lack of adaptability, \ie it considers only \textit{how} to plan shortest paths without \textit{when} to plan it.
%The lack of adaptability makes the planning does not lead to a shorter makespan.

\fakeparagraph{Idea}
%We propose \sysname.
Instead of immediately processing all the racks with item arrival, our \sysname framework only selects a subset of racks for robots to pick up and deliver.
To make this selection be adaptive, \sysname reformulates the problem as a Markov decision process and incorporates a reinforcement learning based rack selection according to the dynamic item throughput.
It also involves a set of efficiency optimizations for both selection and path finding in large-scale warehouse applications.

\fakeparagraph{Workflow}
The workflow of \sysname can be summarized as four steps: \textit{(i)} collect spatiotemporal data from pickers, racks and robots, then \textit{(ii)} use Q-learning to train the model and \textit{(iii)} derive the rack selection scheme $S_t$ containing racks for which \textit{(iv)} it plans paths.

Next, we introduce our adaptive task planning (\secref{sec:atp}) and explain how to improve its efficiency (\secref{sec:efficient}). 

% Our solution to the TSRW problem is \underline{L}earning \underline{T}o \underline{S}chedule (LTS), an efficient reinforcement learning based scheduling framework.
% Its basic idea is to decouple the assignment and planning in the TSRW problem two separate stages: \textit{planning-aware rack selection} followed by \textit{robot path planning}.
% The planning-aware rack selection module optimizes the makespan and generates a series of selection schemes $\{S_i\}$ sequentially considering the dependency between assignment and planning, as well as the online arrival of items.
% The robot path planning module takes each selection $S_i$ as input, and outputs the corresponding planning scheme $U_i$ that satisfies the conflict-free constraints.
% Such designs take advantages of existing conflict-free path finding strategies and shift the scheduling complexity to rack selection, which can be effectively tackled via reinforcement learning.
% To support large-scale warehouse applications, LTS also involves a set of efficiency optimizations for both planning-ware rack selection and robot path planning.
% \figref{fig:system} illustrates the workflow of LTS.
% As next, we elaborate on the design of the planning-aware rack selection and robot path planning module in sequel.

\section{Adaptive Task Planning}
\label{sec:atp}
In this section, we present adaptive task planning.
We model rack selection from the Markov decision process perspective (\secref{subsec:model}), and exploit reinforcement learning for model training and rack selection (\secref{subsec:training}).
For each selected rack, we find conflict-free paths (\secref{subsec:pathfinding}).
At last we integrate rack selection and path finding in \secref{subsec:together}.

\begin{algorithm}[t]
\label{alg:atp}
    \caption{Adaptive Task Planning}
	\Input{$t$: timestamp, $A$ idle robots, $R$: racks , $P$: pickers, $\delta$: bootstrap degree, $\beta$: learning rate, $\epsilon$: policy derivation}
	\Output{$U_t$: planning scheme at $t$}
	\textbf{Initialize} \\
	    \text{initialize } $q$ \\
	    \text{initialize spatiotemporal graph} $G$ \\ 
	\textbf{Rack Selection Step} \\ 
	$\text{approximate} \leftarrow \text{Sample from Bernoulli}(\delta)$ \\
    \If{$\text{approximate} = 1$}{
        $S_t \leftarrow \text{Select same as Naive Task Planning}$ \\
        \For{$ r \in S_t$}{
	        update $q$ by \equref{eq:update} \\
	    }
    }\Else{
         $S_t \leftarrow \emptyset$ \\
         sort $R$ in descending order based on $q(s_r, 0)$ \\
         \For{$r \in R$}{
            $\text{action } \leftarrow \epsilon \text{-greedy}$ \\
            \If{$\text{action} = 1$}{
                $S_t \leftarrow S_t \cup \{\langle r\}$ \\
                update $q$ by \equref{eq:update}, where $c$ is calculated by \equref{eq:reward} \\ 
            }
            \If{$|S_t| = |A|$}{
                \textbf{break} \\
            }
         }
    }
    \textbf{Path Finding Step} \\ 
    $U_t \leftarrow \emptyset$ \\
    \For{$r \in S_t$}{
        $a \leftarrow \text{find the closest robot of } r$ \\
        $u_a \leftarrow \text{find the path on spatiotemporal graph}$ \\
        $U_t \leftarrow U_t \cup \{u_a\} $ \\
        $\text{insert the } u_a \text{ into } G$ \\
    }
	\Return{$U_t$}
\end{algorithm}

\subsection{Markov Decision Process Model}
\label{subsec:model}

The rack selection decisions are made sequentially every timestamp.
It requires considering the rack and its corresponding picker's status (\ie the rack's containing items and the picker's queue and workload and so on).
The selection decisions only depend on the \textit{current} status of racks and pickers instead of \textit{historical} ones, which implies that rack selection decision is sequential decision with Markov property.
Hence, we can derive the definition of rack selection Markov decision process as below.

\fakeparagraph{State}
Since our selection decisions are rack-centric, we define the state of each rack as $\langle ap_r, ar_r \rangle$, where $ap_r$ is the accumulative processing time of the picker associated with rack $r$, \ie $ap_r = \sum_{i=1}^{t}{\mathbb{I}_{p_r\text{ is processing at }i}}$, $ar_r$ is the accumulative processing time of rack $r$, \ie $ar_r = \sum_{i=1}^{t}{\mathbb{I}_{r\text{ is being processed at }i}}$, where $\mathbb{I}_w$ is the indicator function, which is 1 if $w$ is true. 

\fakeparagraph{Action}
The action is also defined from the rack's perspective.
The action $\alpha_r$ becomes requesting pickup, delivery and processing, which is $1$ if $r$ asks for a robot and $0$ otherwise.
The definition dramatically reduces the actions space as binary.
If we define the action from meta view and the action is to directly select racks, the action space would be combinatorial and difficult to cope with.

\fakeparagraph{State Transition}
Based on the definition of action $\alpha$, the transitions are as below.
If $\alpha_r = 0$, the state remain the same.
If $\alpha_r = 1$, state $\langle ap_r, ar_r \rangle$ will transit to $\langle ap_r + \sum_{i \in \tau_r}i, ar_r + \sum_{i \in \tau_r}i \rangle$.
That is, the accumulative processing time of both the picker and the rack will increase by the total processing time of the current items on rack $r$, \ie $\sum_{i \in \tau_r}i$.

Though a single transition always changes $ap_r$ and $ar_r$ in the same way, updates from different racks can make $ap_r$ and $ar_r$ different since one picker can be associated with multiple racks.
The fact that one picker is responsible for multiple racks also implies the dependence among racks, which motivates the joint state modeling of the rack and its picker.

\fakeparagraph{Reward}
We use negation of the increment in a picker's finish time $f_p$ after selecting certain rack as the reward.
However, it is difficult to derive the increment of picker's finish time.
This is because the rack selection decision is performed before path planning, and thus the delays for pickup, delivery, queuing and return are unknown.
Thus we estimate the reward as follows.
\begin{equation}\label{eq:reward}
    c =  - \left(\max\{f_p, d(l_r, l_{p_r})\} + \sum_{i \in \tau_r}i\right)
\end{equation}
where the max term denotes the increase in waiting time and the sum term is the increment of $ap_r$.
The negation is because our goal is to minimize makespan while reinforcement learning maximizes the sum of rewards.

Note that our reward design considers the end-to-end delay for pickup, delivery, queuing, processing, and return, and thus is aligned with our makespan definition.

\fakeparagraph{Optimizations}
Based on the above Markov decision process definition, optimizing makespan requires us to find \textit{policy} that accounts for the current states and make actions for all racks.
The policy is derived from \textit{value function}, which is represented as $q(\langle ap_r, ar_r \rangle, \alpha)$ in our problem.
It maps the state-action to the expected accumulative rewards.
Based on the definition of reward, the value function indicates the expected finishing time of rack $r$ considering both its delivery time and the picker's finish time.
According to the value function $q$, the best action would be $\arg\max_{\alpha'}q(s, \alpha')$.
However, this policy may be trapped into sub-optimal solutions because $q(s, \alpha)$ can be inaccurate especially in the early stage of training.
Instead, we adopt the $\epsilon$-greedy policy \cite{sutton2018reinforcement}.
It chooses the current best action with $1 - \epsilon$ probability and a random action with $\epsilon$ probability to balance exploration and exploitation.

\begin{figure}[t]
    \centering
		\includegraphics[width=0.8\linewidth]{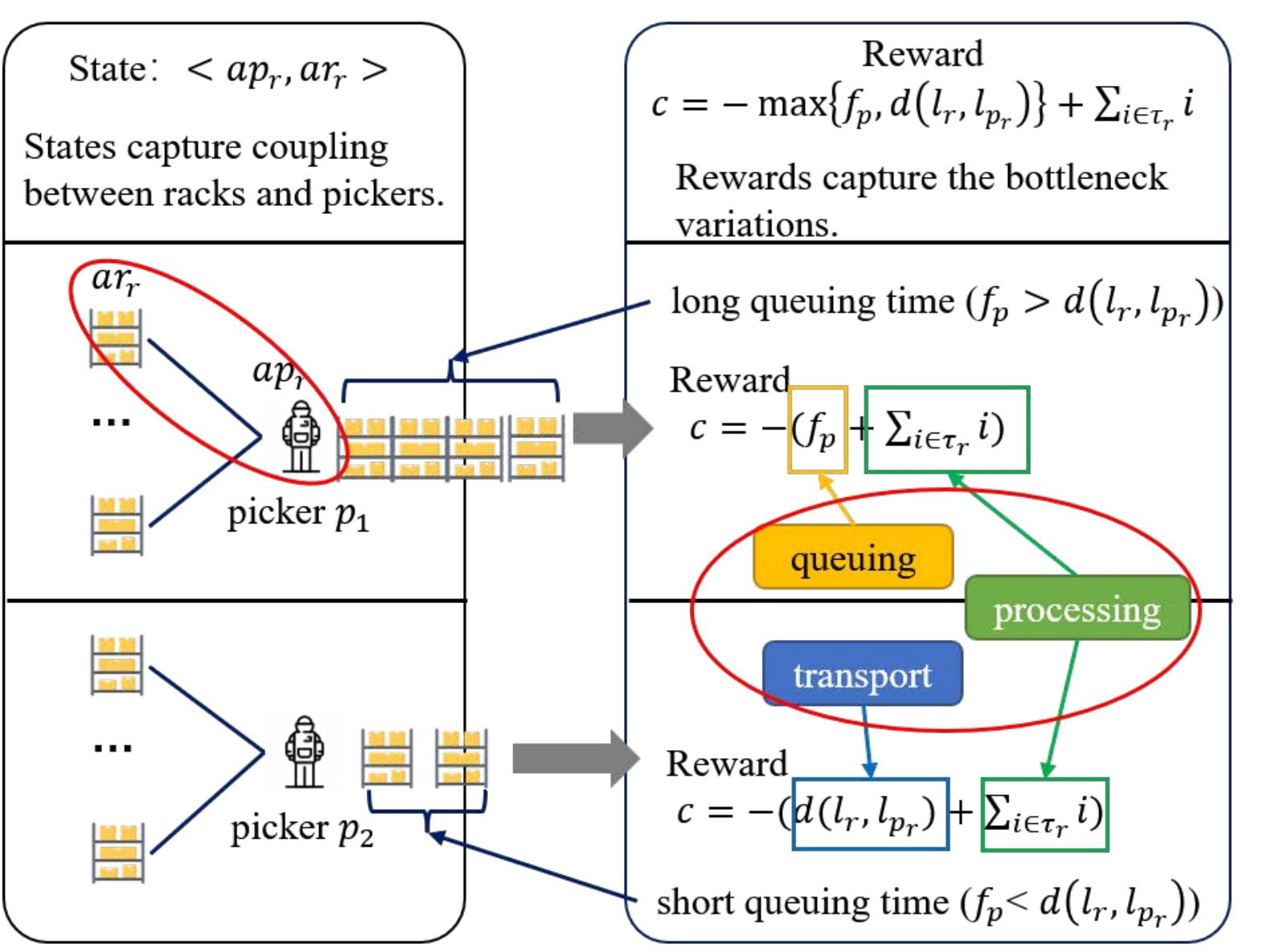}
    \caption{Illustration of our Markov decision process model}
    \label{fig:modelintuitive}
\end{figure}

\fakeparagraph{Remarks}
\figref{fig:modelintuitive} illustrates why our model is fit for the TPRW problem. 
The state definition jointly implies both the rack and its picker, the reward will change while the fulfillment bottleneck varies.

We can derive a policy only if the value function can be effectively trained, as explained below.

\subsection{Q-Learning Based Model Training}
\label{subsec:training}

We apply Q-learning \cite{watkins1989}, a classic temporal difference based bootstrap method to train the value function, since it is better fit for online learning and highly self-adaptive \cite{sutton2018reinforcement}.
The Q-learning trains value function as below.
\begin{equation}\label{eq:update}
    q(s, \alpha) \leftarrow  q(s, \alpha) + \beta \cdot (c + \gamma \max_{\alpha'}q(s', \alpha') - q(s, \alpha))
\end{equation}
where $s$/$s'$ are the current/next state after taking action $\alpha$, $\beta$ is the learning rate, $c$ is the reward and $\gamma$ is the discount factor.

Directly applying Q-learning as above will bootstrap \textit{unexplored} states. 
Recall that state $\langle ap_r, ar_r \rangle$ is time-dependent.
When updating the value function by \equref{eq:update}, $s'$ is $\langle ap_r + \sum_{i \in \tau_r}i, ar_r + \sum_{i \in \tau_r}i \rangle$ (see state definition in \secref{subsec:model}).
The new state is unexplored because both $ap_r$ and $ar_r$ always increase, preventing the value function from converging.
As a remedy, we integrate a greedy method into the training.
Specifically, at each timestamp, we choose the greedy method with probability $\delta$ and the original bootstrap with $1 - \delta$ probability.
This way, the greedy method will provide solutions, explore some states and update the value function approximately.
Then based on the approximation, bootstrap is able to train the value function more precisely.
The greedy method adapts the ``most slack picker first'' strategy.
That is, it greedily chooses those racks whose associated picker has the smallest $f_p$.

\begin{figure}[h]
    \centering
    	\subfigure[Spatial graph]{
		\includegraphics[width=0.25\linewidth]{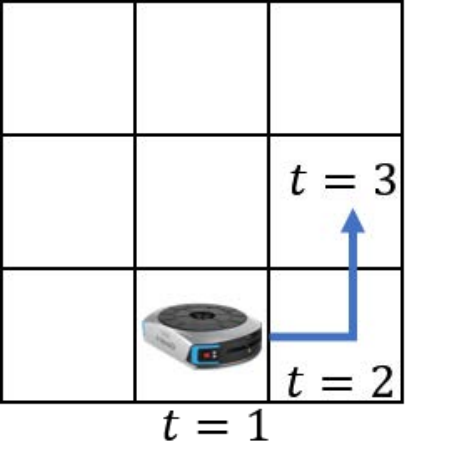}
		\label{fig:sg}
	}
	\subfigure[Spatiotemporal graph]{
		\includegraphics[width=0.4\linewidth]{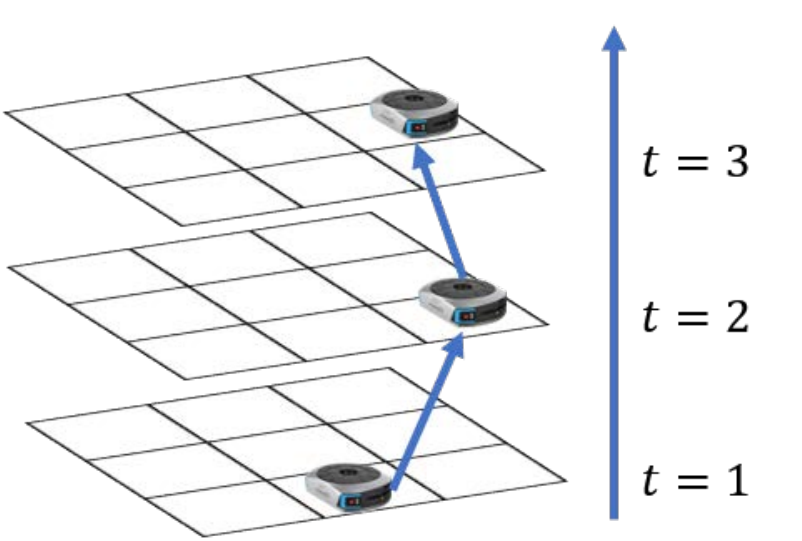}
		\label{fig:stg}
	}
    \caption{Illustration of a spatial graph and a spatiotemporal graph.}
    \label{fig:graph}
\end{figure}

\subsection{Path Finding on Spatiotemporal Graph}
\label{subsec:pathfinding}

Given the selection schemes, a path finding algorithm plans conflict-free paths.
We adopt A* algorithm \cite{TSSC68Peter} for finding conflict-free paths.
Instead of searching on the spatial graph, the algorithm searches on the spatiotemporal graph to avoid conflicts.
Intuitively, the space graph is duplicated in every time step.
Each vertex represents a location with a certain timestamp, while each edge represents two vertex are adjacent both spatially and temporally (see \figref{fig:graph}) \cite{AAMAS16Ma}.
On spatiotempoal graph, the algorithm starts with the source vertex (\ie grid with certain timestamp) and then maintain a open set which stores vertices explored.
Based on the current cost and heuristic value (h-value), we choose a vertex from the open set for the next round search.
When searching on the grid based space, the h-value is usually defined as the Manhattan distance from current vertex to the destination \cite{ICAPS19Florian}.

\subsection{Put it Together}
\label{subsec:together}
By integrating rack selection and path finding, we propose \underline{A}daptive \underline{T}ask \underline{P}lanning (ATP), as shown in  \algref{alg:atp}.

Lines 4 to 19 are rack selection step while lines 20 to 26 are path finding step.
In rack selection step, we first randomly decides whether to approximate or bootstrap (line 5).
If approximate, we use the greedy method to derive the selection scheme (lines 6 to 9).
If bootstrap, we sort racks based on the value function in line 12.
It will then preferentially select racks with the largest expected finish time till no robot is available (line 13 to 19).
Both steps adopt Q-learning to update the value function (line 9 and 17).
Based on selection scheme calculated from selection step, path finding step will assign the closest robot of each selected racks (line 23) and finds path for it (line 24).
The spatiotemporal graph will maintain all prior planed path for conflict avoidance (line 26).
Note that $\delta$ controls the degree of bootstrap.
A larger $\delta$ means smaller bootstrap.
From empirical results (see \secref{subsec:expResults}), a $\delta$ smaller than 0.4 contributes to effective training. 

% \fakeparagraph{Time Complexity}
% The complexity of the rack selection step is $O(|R|\log|R|)$.
% For approximate operations, the complexity is $O(|P|\log|P| + |A|)$, where the two terms correspond to sorting and selecting (lines 6 to 9 in \algref{alg:atp}).
% The bootstrap operations take $O(|R|\log |R| + |A|)$ time, which corresponds to sorting (line 10 in \algref{alg:atp}) and selecting (line 11 to 17 in \algref{alg:atp}).
% So, the expected complexity of rack selection is $O(\delta \cdot (|P|\log|P| + |A|)  + (1 - \delta) \cdot (|R|\log|R| + |A|) = O(\delta |P|\log|P| + (1-\delta) |R|\log|R| + |A|)$.
% Considering $|P| << |R|$, $|A| < |R|$, and $\delta$ is set to smaller than 0.4, the complexity is $O(|R|\log|R|)$.

% The complexity of the path finding step is $O(|R|(HW)^2)$, where $|R|$ is the number of racks and $H$/$W$ is height/width of of the warehouse.
% This is because the path finding conduct A* algorithm for all selected racks.

\begin{figure*}[t]
    \centering
    \includegraphics[width=0.9\linewidth]{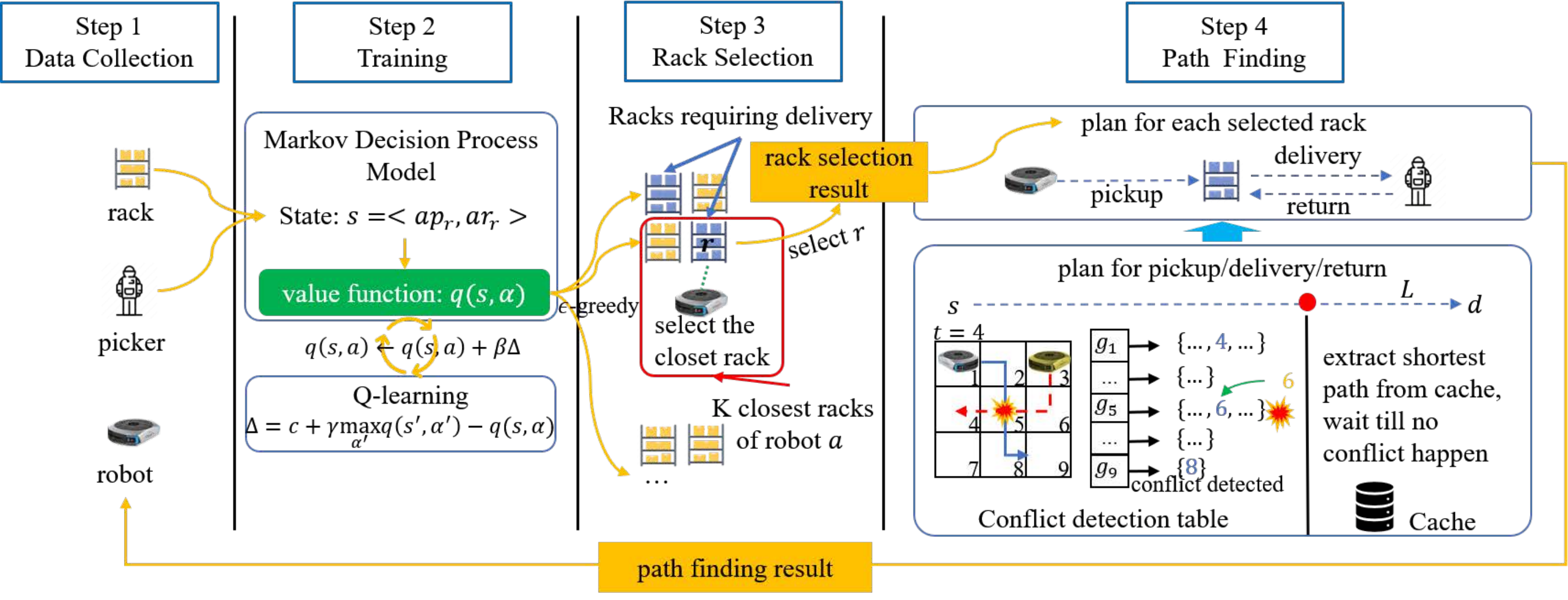}
    \caption{Workflow of Efficient Adaptive Task Planning (\sysname).}
    \label{fig:workflow}
\end{figure*}

\section{Efficiency Optimizations for Adaptive Task Planning}
\label{sec:efficient}

In this section, we present our efficient design for adaptive task planning.
For rack selection step, we flip the requesting side from rack to robot to reduce selection time consumption (\secref{subsec:effSelection}).
For path finding step, we optimize both the time and memory consumption.
We replace the spatiotemporal graph with conflict detection table which has less space complexity while support for quick conflict detection and use cache aiding the finding step (\secref{subsec:effFinding}).
% (\secref{subsec:table}).
% For accelerating the path finding step, we use cache aided the finding step (\secref{subsec:cache}).
At last we elaborate integrate these efficient design into ATP (\secref{subsec:integrate}).

\subsection{Optimization for Rack Selection}
\label{subsec:effSelection}

From the time complexity analysis above, the bottleneck of ATP's rack selection step lies in sorting racks.
Instead of traversing racks then requesting robot for delivery, we accelerate the process by flip requesting side to robot.
Specifically, we traverse robots and then finding a rack among its closest $K$ racks.
Since all racks' locations in the storage area are fixed, recording closest $K$ racks of different grids is static and easy to maintain.
Then for each robot, we can easily find its closest racks according to its located grid and find the closest selected rack among those racks (See \figref{fig:workflow}).

% \begin{figure}[htb]
%     \centering
%     \includegraphics[width=0.6\linewidth]{fig/flip.pdf}
%     \caption{Illustration of flip requesting side.}
%     \label{fig:flip}
% \end{figure}

\subsection{Optimization for Path Finding}
% \label{subsec:table}
\label{subsec:effFinding}

\fakeparagraph{Memory Compression via Conflict Detection Table}
Searching on a spatiotemporal graph incurs large space cost due to the ever-increasing temporal dimension.
In worst case, the space complexity of the spatiotemporal graph is $O((HW)^2)$ because the space complexities of the spatial graph and the path lengths are both $O(HW)$.
Next we introduce our conflict detection table to reduce the space complexity while maintaining efficient conflict detection.

% \begin{figure}[htb]
%     \centering
%     \includegraphics[width=0.8\linewidth]{fig/exp_table.pdf}
%     \caption{An example for conflict detection table. Each grid contains all trajectories passing time. When planning a new path, the conflict can be quickly detected.}
%     \label{fig:expTable}
% \end{figure}

In conflict detection table, an array is built for all grids, and each entry contains a set recording the passing time.
The set can either be implemented based on balanced binary search tree or hash set for quick search.
When planning path, ATP will search for a grid and by checking the grid's corresponding entry contains the timestamp or not, it will quickly judge whether conflict will happen.
This conflict detection table removes the ever growing temporal dimension.
The space complexity is decreased to $O(HW)$.

In addition to path finding, the table also supports update and insertion.
The update operation deletes all passed timestamp.
This operation reduces the space cost of the table and is executed periodically.
The insertion operation inserts a path to the table.
It will insert the passing time to the corresponding grid entry for each point of trajectory.

% \subsection{Acceleration for Path finding via Cache}
% \label{subsec:cache}

\fakeparagraph{Cache-aided Path Finding}
We can further accelerate the path finding algorithm by caching certain shortest paths without considering conflict, and then deriving the conflict-free paths based on the shortest paths.

As mentioned in \secref{subsec:pathfinding}, the path finding algorithm will maintain a open set and pick vertex from it for path finding.
Specifically, when picking vertex from the open set, if the distance between the current vertex and the destination is within a threshold $L$, we directly extract the shortest path from the cache and derive the conflict-free path.
The corresponding policy is to let the robot wait till there is no conflict to move next steps along the shortest path.
The rationale is that, when the current vertex is close to the destination (\ie within the threshold), instead of searching for the shortest conflict-free path, directly moving along the shortest path with some wait may be effective, since it is already close to the destination.
Such cache can notably reduce the size of open set and thus the search cost.

\subsection{Integrate Efficient Design into ATP}
\label{subsec:integrate}

Integrating all efficient designs, we get our final \underline{E}fficient \underline{A}daptive \underline{T}ask \underline{P}lanning (EATP), as in \algref{alg:eatp} and \figref{fig:workflow}.

It first initializes the cache (line 3) and conflict detection table (line 4).
The cache contains shortest paths whose Manhattan distances are within threshold $L$.
For rack selection, lines 10 to 13 are flip requesting and line 18 is cache-aided path finding.
It will find paths by CDT and when the current vertex is close to the destination ($l_r$ or $l_p$ depended on different steps of pickup, delivery or return) less than $L$, it will derive the last segment of path by waiting till no conflict occurs.

For the hyperparameter distance threshold $L$, it controls the degree of cache-aiding, a larger value encourages using cache for less computation consumption.

\begin{algorithm}
\label{alg:eatp}
    \caption{Efficient Adaptive Task Planning}
	\Input{$t$: current timestamp, $A$: idle robots, $P$: pickers, $R$: racks, $L$: distance threshold}
	\Output{$U_t$: planning scheme at time $t$}
	\textbf{Initialize} \\
	\text{initialize } $q$ \\
	\text{initialize } $Cache$ containing all shortest paths with length $\le L$. \\
	\text{ initialize conflict detection table } $CDT$ \\
	\textbf{Rack Selection Step} \\ 
	$\text{approximate} \leftarrow \text{Sample from Bernoulli}(\delta)$ \\
    \If{$\text{approximate} = 1$}{
        $S_t \leftarrow \text{Same as lines 7 to 9 in \algref{alg:atp}}$ \\
    }\Else{
         \For{$a \in A$}{
            \For{$r \in \{K \text{ racks closet to }l_a \}$}{
                $\text{Update } S_t, q\text{ same as line 14 to 17 in \algref{alg:atp}}$ \\ 
                \text{break inner loop when a rack is selected}
            }
        }
    }
	\textbf{Path Finding Step} \\ 
	$U_t \leftarrow \emptyset $ \\
	\For{$r \in S_t$}{
	    $a \leftarrow \text{ find closet robot to } r$ \\
	    $u_a \leftarrow \text{path finding on } CDT \text{ and derive via } Cache$ \\
	    $U_t \leftarrow U_t \cup \{u_a\}$ \\
	    $CDT.insert(u_a)$ \\
	}
	\Return{$U_t$}
\end{algorithm}

% \fakeparagraph{Time Complexity}
% For rack selection step, with the flip selection, the time complexity is $O(\delta (|P|\log|P| + |A|) + (1 - \delta) (K|A|))$ corresponding to approximate and bootstrap operations.
% Since $\delta$ is small, the complexity is $O(K|A|)$.
% By flip requesting side, we reduce the complexity of rack selection from $O(|R|\log|R|)$ to $O(K|A|)$.
% The number of racks in a warehouse is much greater than that of robots so this reversing design will largely reduce the time consumption.
% In \secref{subsec:expResults} we will show that this design will accelerate the execution speed by up to 280\%.

% As for path finding step, our cache will help avoid searching the complete path thus largely reduces the size of the open set.
% In \secref{subsec:expResults} we will show that this design will reduce planning time consumption by mostly 75.5\%.
\section{Evaluation}
\label{sec:evaluation}
This section presents the evaluations together with case study of our proposed methods.

\subsection{Experimental Setup}
\label{subsec:settings}

\fakeparagraph{Datasets}
We use both synthesized and real datasets.
The two synthesized datasets Syn-A and Syn-B are generated on two warehouse layouts with different numbers of items.
All items emerge following Poisson distribution and each racks picking time is distributed uniformly between 20 and 40 seconds, which is close to the real situation.
Two real datasets are derived based on historical records from Geekplus, one of the world's leading smart logistics companies \footnote{https://geekplusrobotics.borealtech.com/en/ \label{geek}}.
Two real datasets are named as Real-Normal and Real-Large considering the scalability of data.
\tabref{tab:dataset} lists the dataset details.

\begin{table}[t]
	\centering
	\caption{Summary of datasets.}
	\label{tab:dataset}
	\small
	\begin{tabular}{ccccc}
	    \toprule
		Name    &$H \times W$ & \#Item & \#Robot  & \#Rack	\\	\midrule
% 		Normal & $233\times104$ & $10^5$ & $408$ & $4896$ \\
% 		Large & $426\times146$ & $5\times 10^5$ & $1056$ & $12672$\\
% 		Geekplus1 &$240\times206$ & $557128$ & $952$ & $9792$ \\ 
% 		Geekplus2 & $541\times302$ & $10^6$ & $2800$ & $33600$  \\
		Syn-1 & $233\times104$ & $ 10^5$ & $0.5$k & $5.0$k \\
		Syn-2 & $426\times146$ & $5\times 10^5$ & $1.0$k & $1.3$k\\
		Real-Norm\textsuperscript{\ref{geek}} & $240\times206$ & $5.6\times 10^5$ & $1.0$k & $10$k \\ 
		Real-Large\textsuperscript{\ref{geek}} & $541\times302$ & $10^6$ & $3.0$k & $34$k  \\
		\bottomrule
	\end{tabular}
\end{table}

\begin{figure}[htp]
    \centering
    \subfigure[Virtual system]{
		\includegraphics[width=0.45\linewidth]{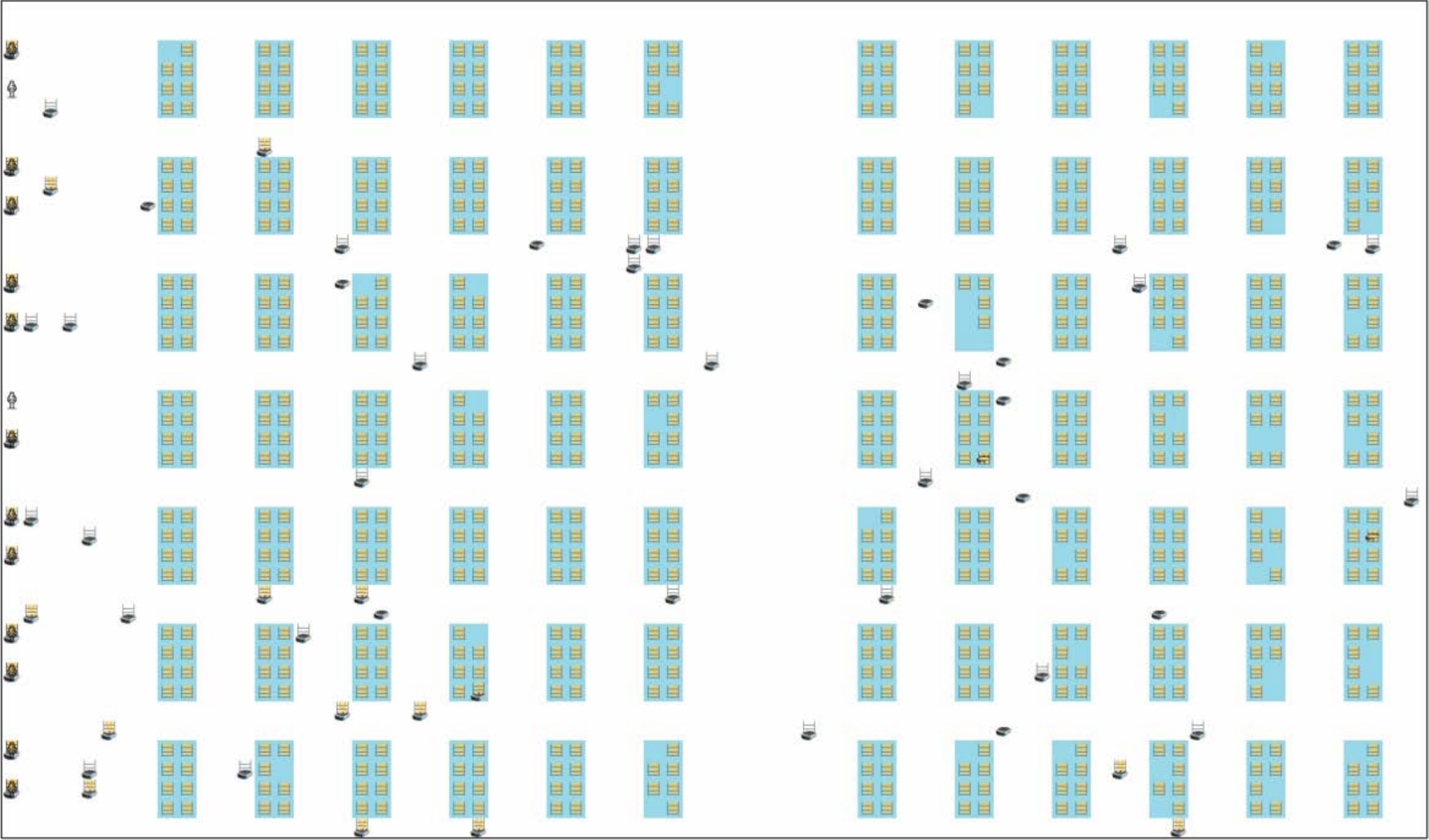}
		\label{fig:virtual}
	}
	\subfigure[Real warehouse]{
		\includegraphics[width=0.45\linewidth]{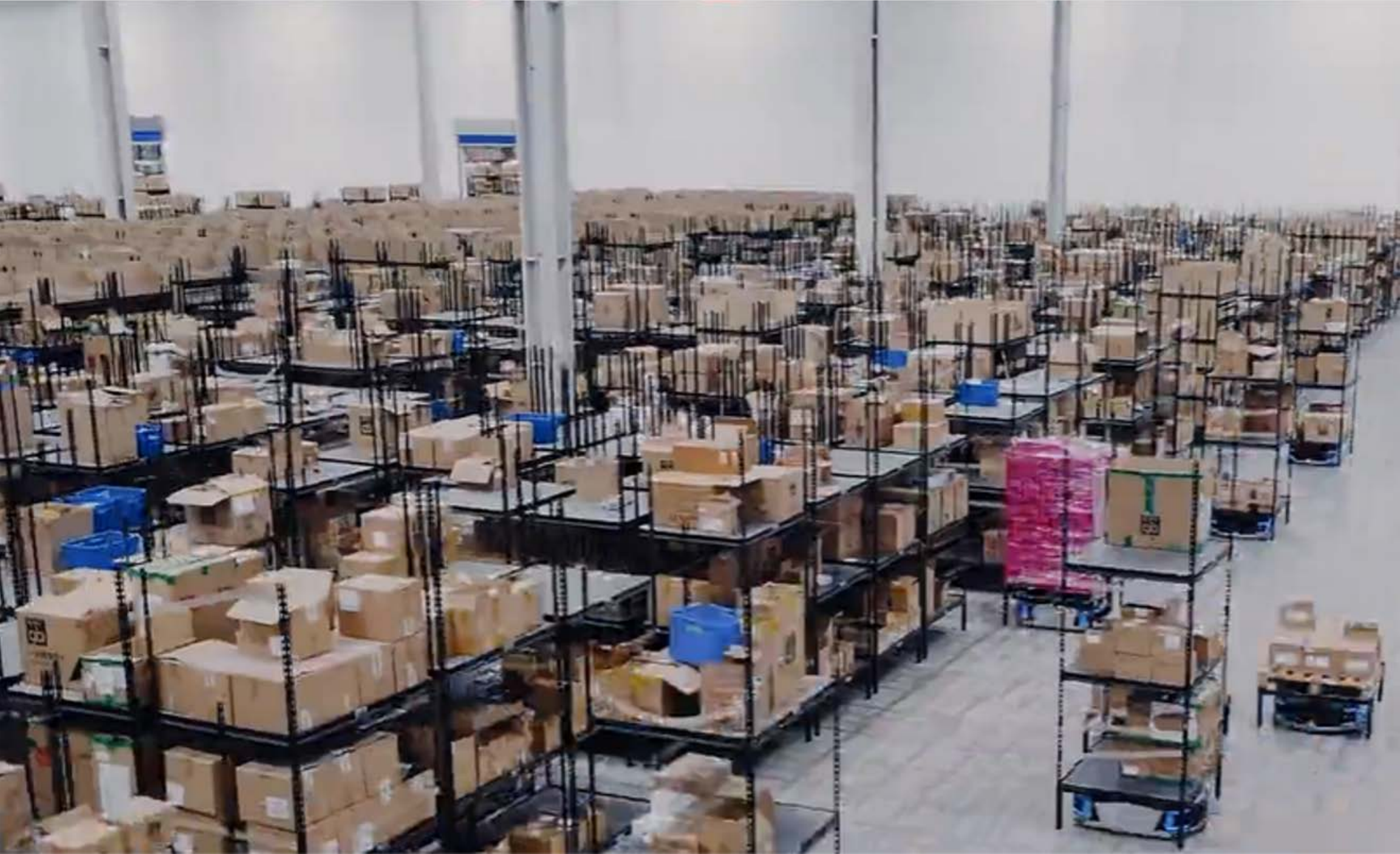}
		\label{fig:expWarehouse}
	}
	\caption{Snapshots of virtual (a) and real (b) warehouses.}
	\label{fig:validation}
\end{figure}

\fakeparagraph{Validation system}
To test algorithms' performances on these datasets, 
We build a virtual warehouse which simulates the movement of robots and the processing of pickers.
At each timestamp, it collects all idle robots and racks containing remaining items as well as pickers' working status, then executes the algorithm for path planning.
Then it converts the path planning scheme to instructions on robots' motion.
It also records the performance of task planning algorithms in terms of effectiveness and efficiency.
A snapshot of the validation system is shown in \figref{fig:virtual}.
A real warehouse is also used for deployment demonstration (See \figref{fig:expWarehouse}).

As for implementation, we set the default value of $\delta$, $\epsilon$, learning rate $\beta$ and distance threshold $L$ to be 0.2, 0.1, 0.1 and 50, respectively.
% Note that we also adopt discretizing tricks splitting timestamp by 100 units a step for state representation, thus the state number will be reduced to $10^4$, much less than original $10^{12}$ (one second as a unit and a whole day as horizon, $ap_r$ and $ar_r$ will both be up to $10^6$).
All algorithms as well as the validation system are implemented in Java 8 and the experiments are run on 4 cores CPU Intel(R) Xeon(R) Platinum 8269CY CPU T 3.10GHz with 20 GiB Java virtual machine memory.

\begin{figure*}[!t]
\centering
	\subfigure[PPR on Syn-A]{
		\includegraphics[width=0.23\linewidth]{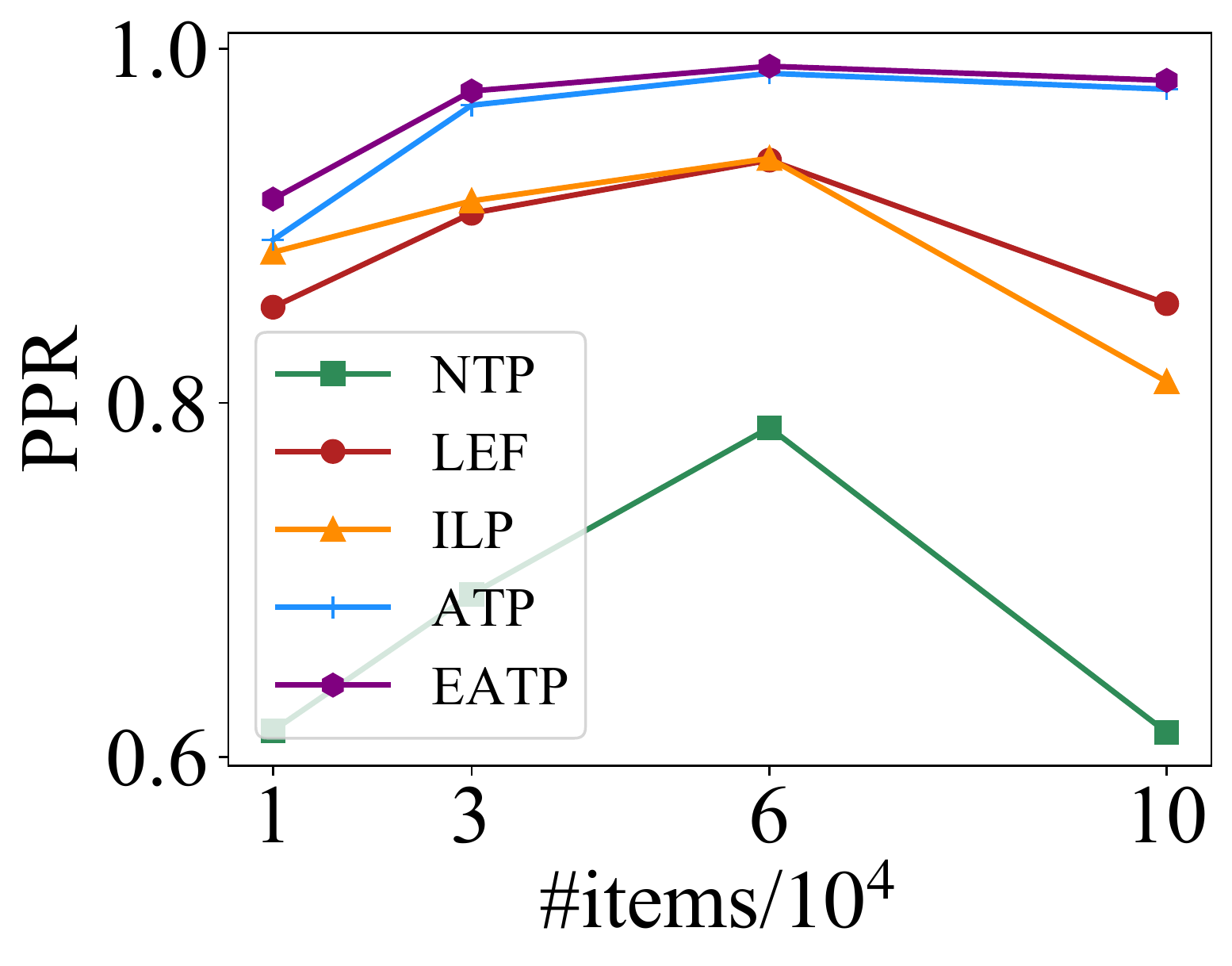}
		\label{subfig:normalPPR}
	}
	\subfigure[PPR on Syn-B]{
		\includegraphics[width=0.23\linewidth]{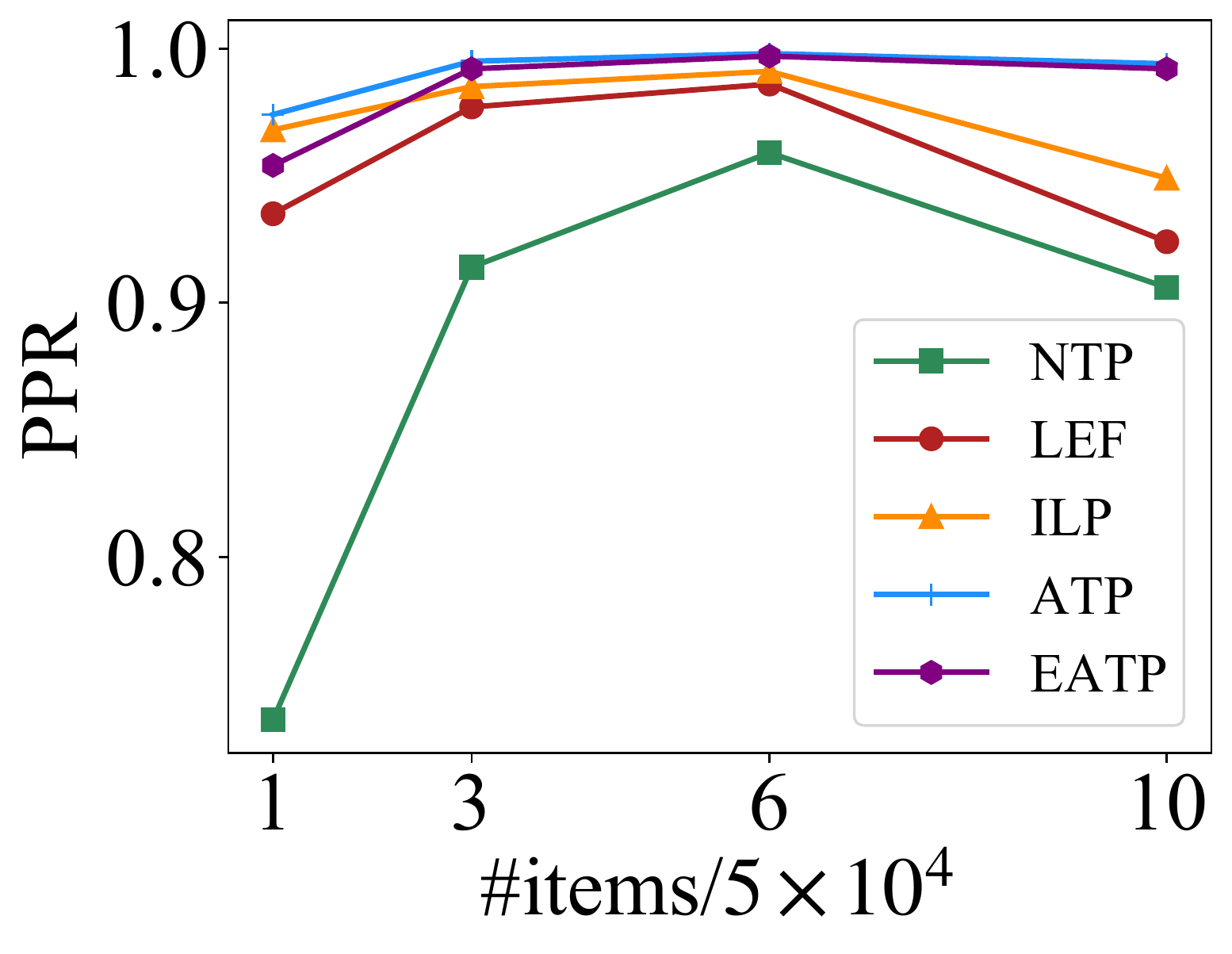}
		\label{subfig:largePPR}
	}
	\subfigure[PPR on Real-Norm]{
		\includegraphics[width=0.23\linewidth]{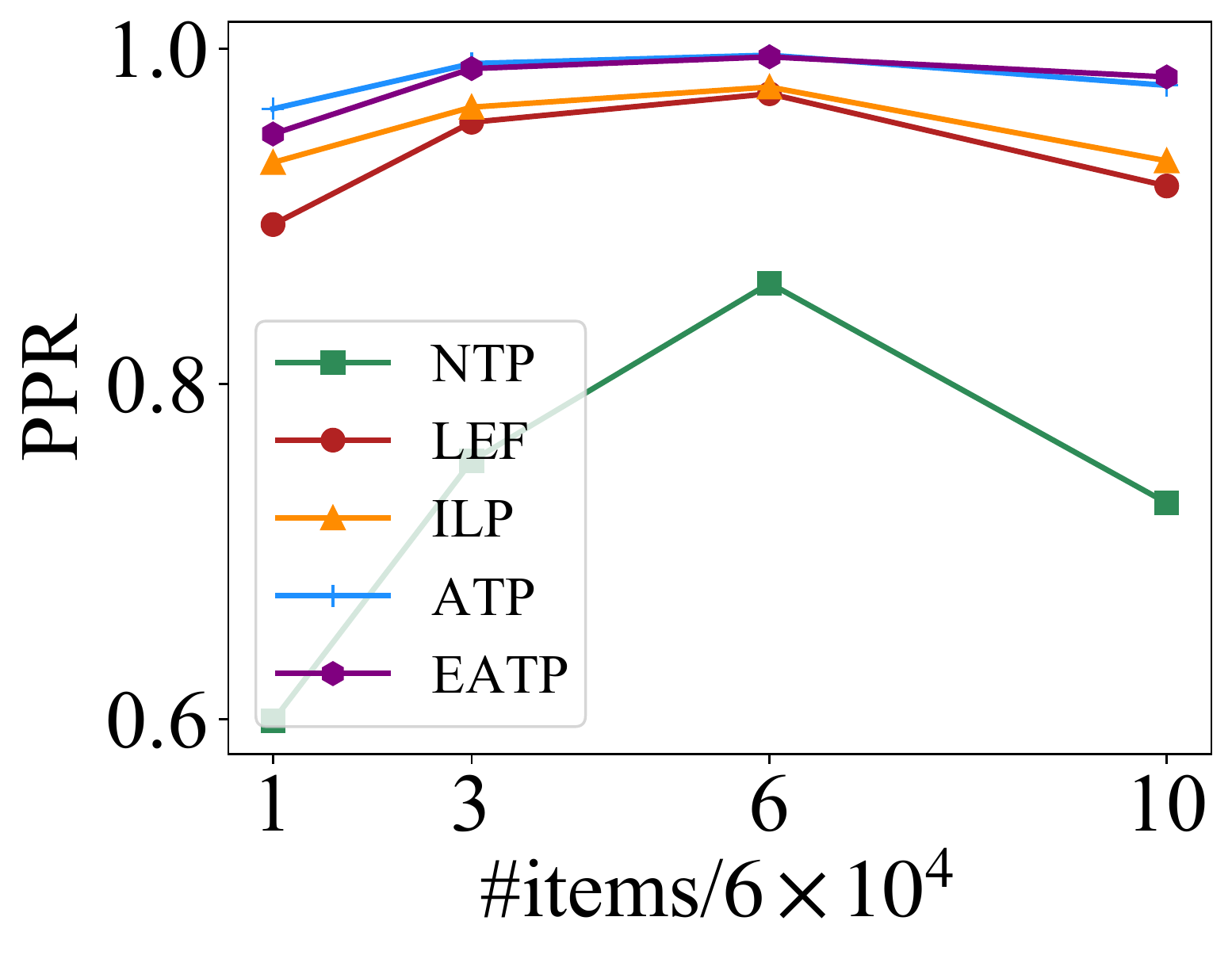}
		\label{subfig:realPPR}
	}
	\subfigure[PPR on Real-Large]{
		\includegraphics[width=0.23\linewidth]{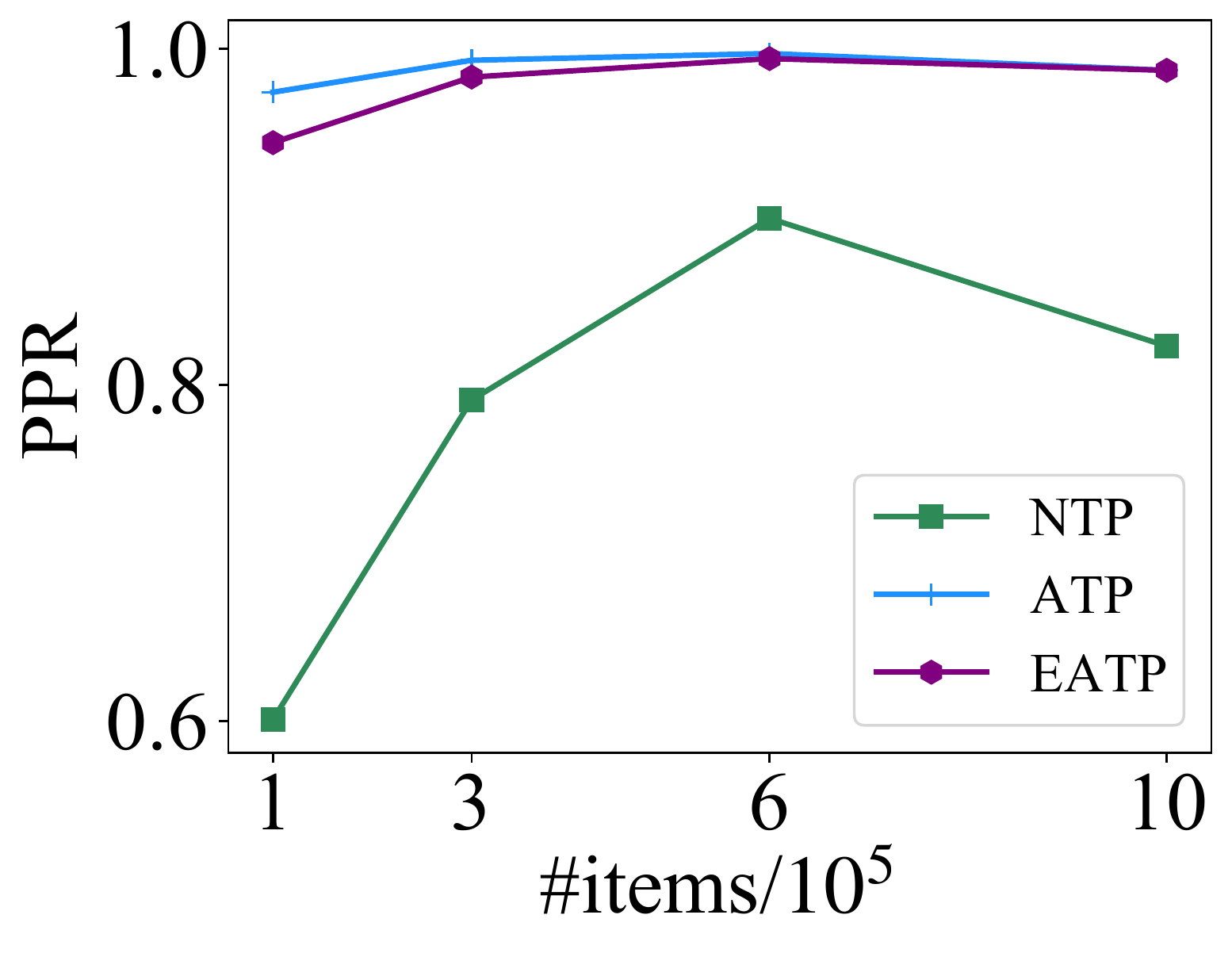}
		\label{subfig:superPPR}
	}
	\subfigure[RWR on Syn-A]{
		\includegraphics[width=0.23\linewidth]{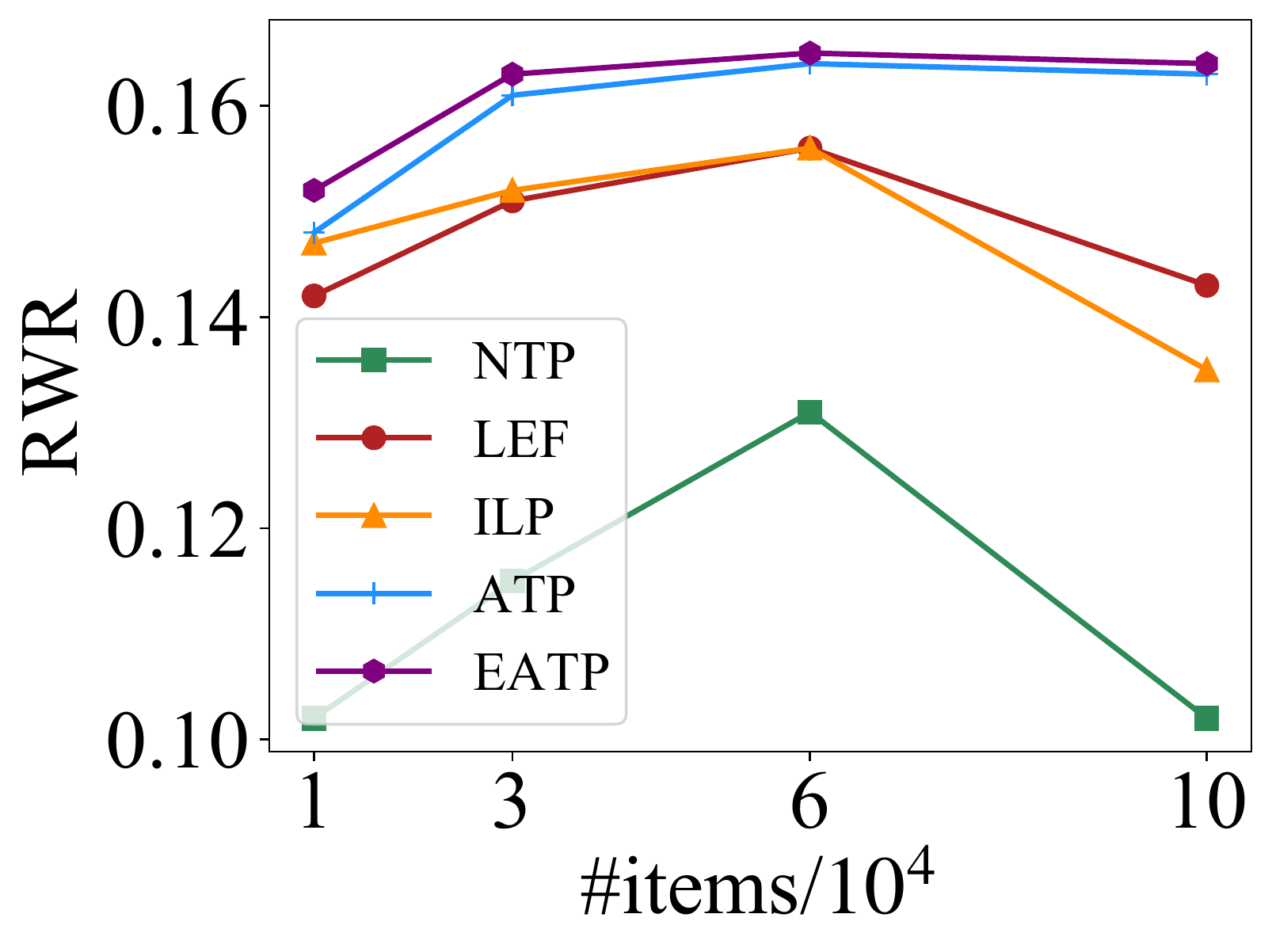}
		\label{subfig:normalRWR}
	}
	\subfigure[RWR on Syn-B]{
		\includegraphics[width=0.23\linewidth]{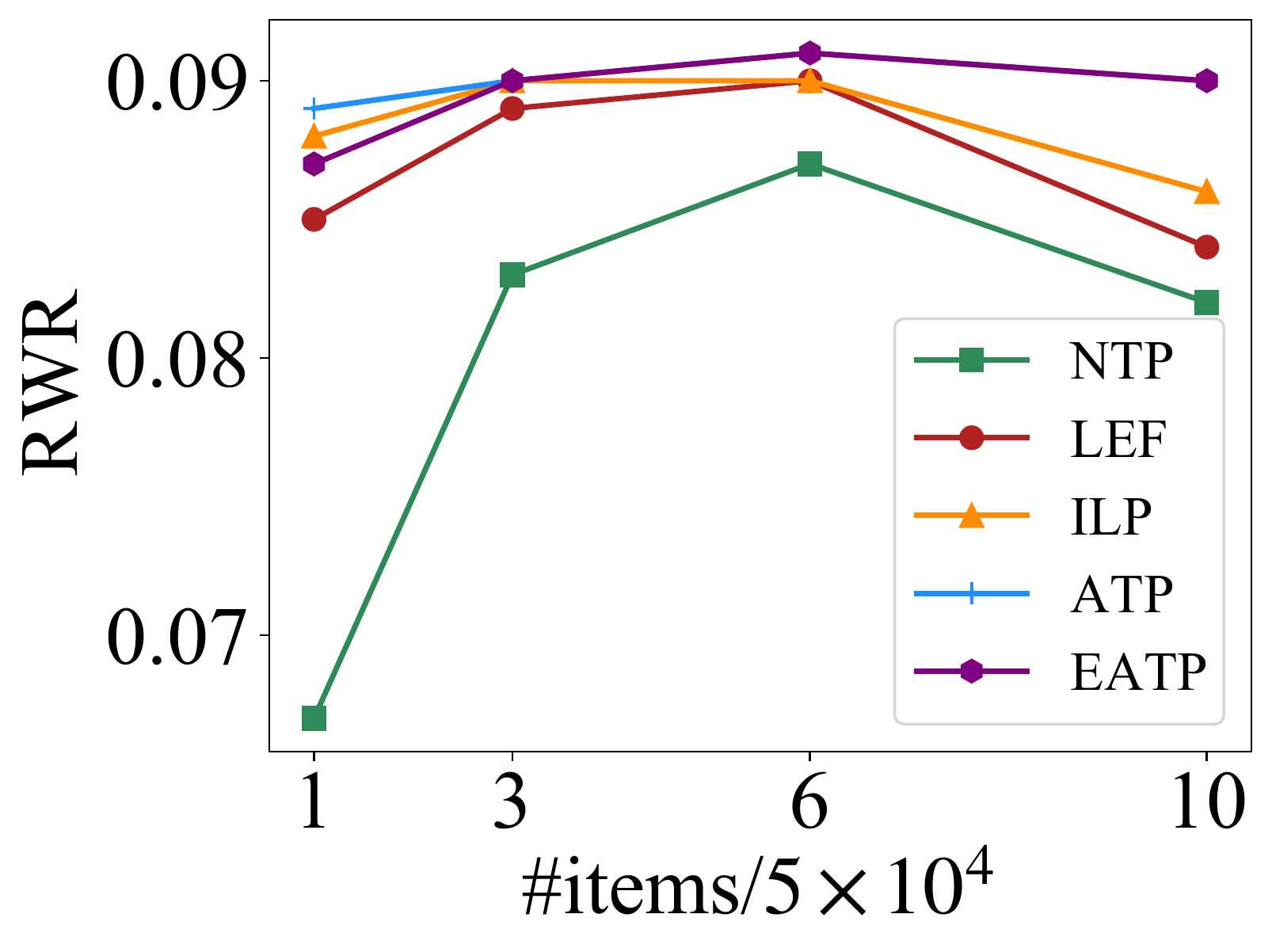}
		\label{subfig:largeRWR}
	}
	\subfigure[RWR on Real-Norm]{
		\includegraphics[width=0.23\linewidth]{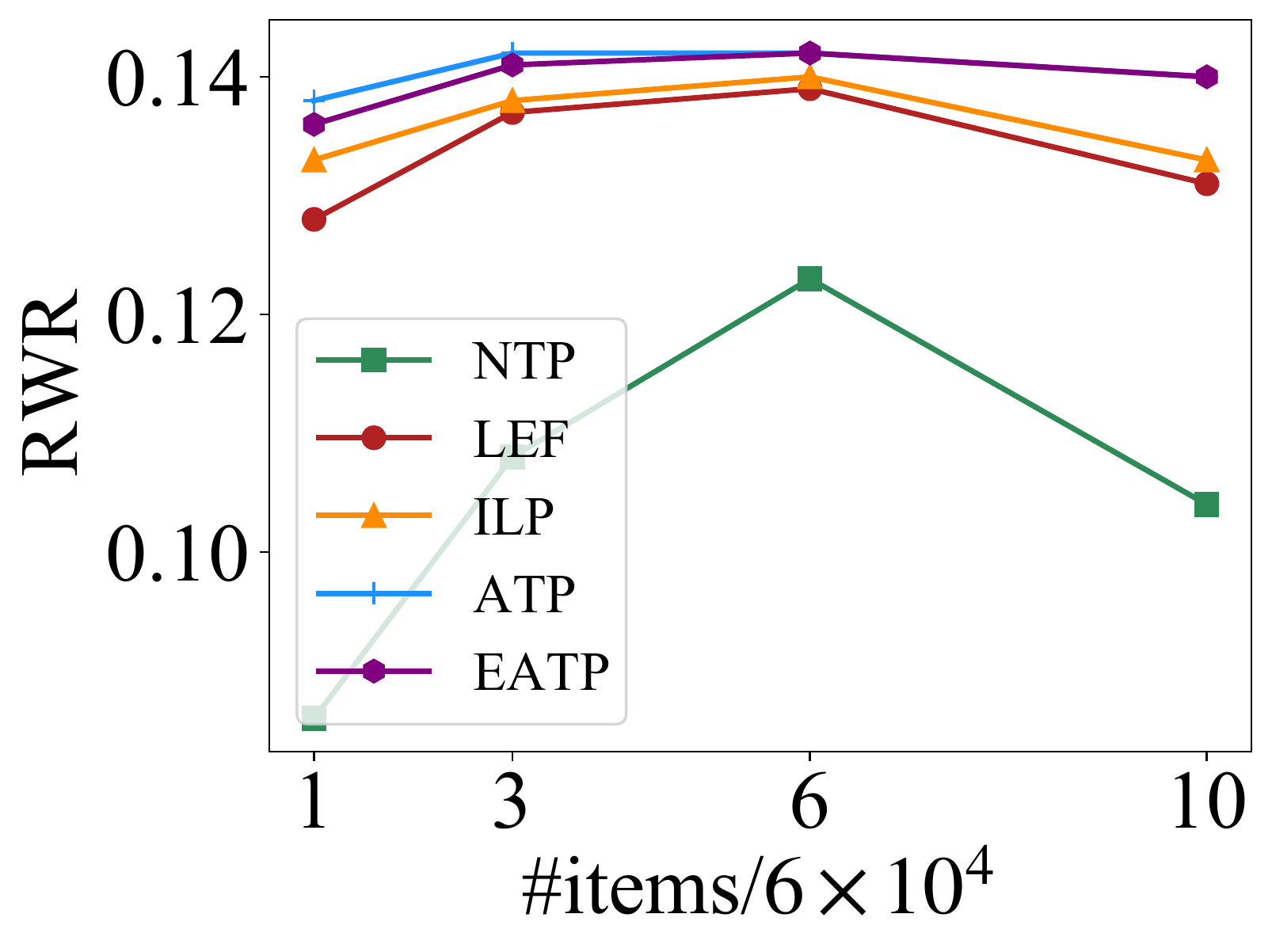}
		\label{subfig:realRWR}
	}
	\subfigure[RWR on Real-Large]{
		\includegraphics[width=0.23\linewidth]{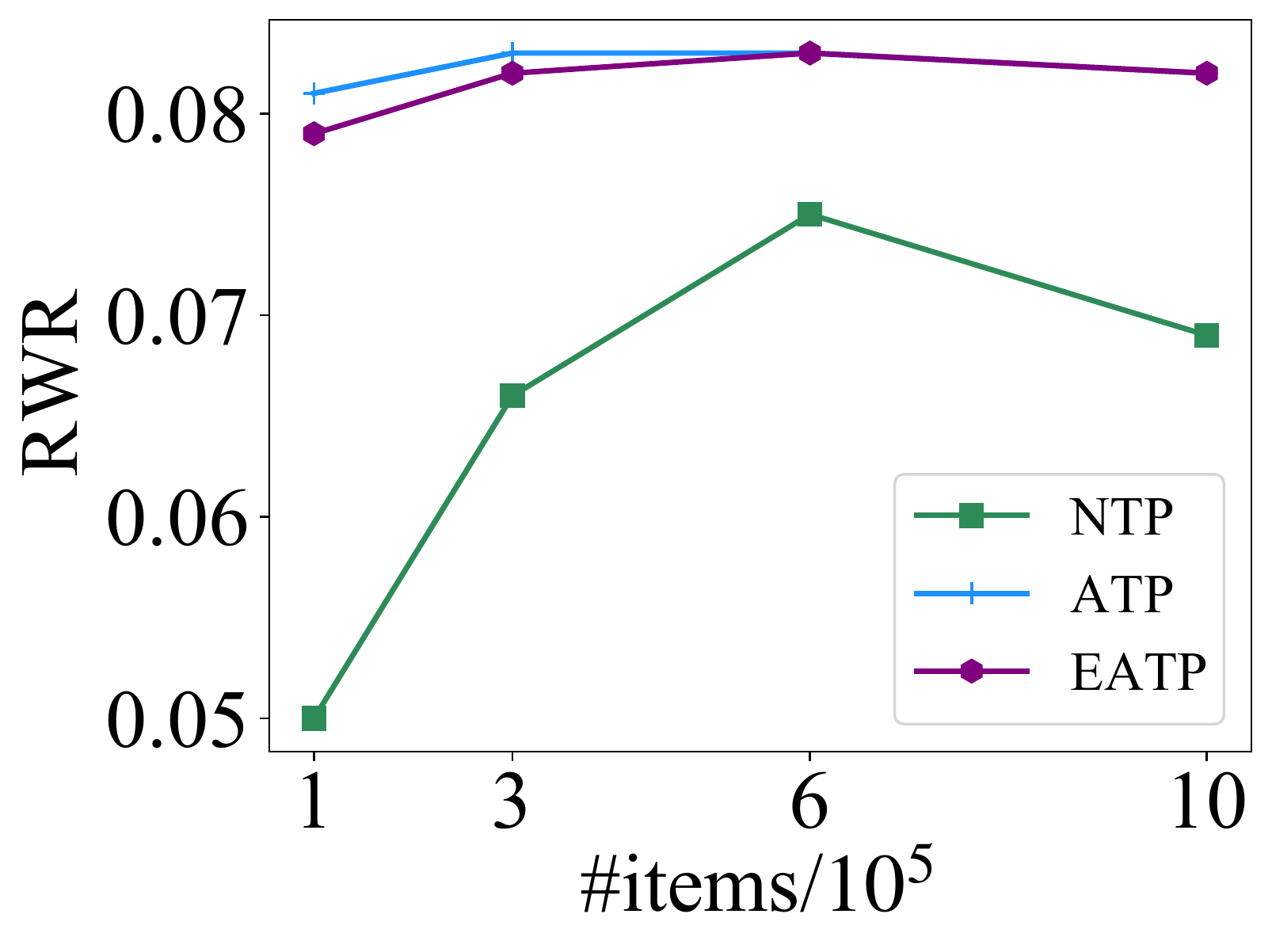}
		\label{subfig:superRWR}
	}
	\caption{Picker's Processing Rate (PPR) and Robot's Working Rate (RWR) comparisons.}
	\label{fig:pprrwr}
\end{figure*}

\fakeparagraph{Baselines} We compare \sysname with the following methods.
% We adopt Token Passing planning algorithm \cite{AAMAS17Ma}, the state-of-the-art method to online multi-agent path planning problem as baseline.
% It finds conflict-free paths in a sequential manner.
% A token is passing among all robots and the one who get the passing try to find its closest task and search a path via A* algorithm.
% Other prior studies \cite{AAAI19vancara, AAAI21Li} assume tasks and robots emerge in a binding manner and is not suitable for our problem. 
% To make the algorithm more adaptive for our setting, we implement below counterparts for \textit{rack selection}.
\begin{itemize}
    \item 
    \textbf{Naive Task Planning (NTP) \cite{AAMAS17Ma}}.
    This method is the directly extension of state-of-the-art path planning algorithm \cite{AAMAS17Ma}.
    It assigns robots to racks whose corresponding picker has the earliest finish time $f_p$ (see \secref{sec:baseline}).
    \item \textbf{Least Expiration First planning (LEF) \cite{GEO16Deng}}.
    This spatiotemporal task selection algorithm selects tasks with least expiration time \cite{GEO16Deng}.
    Though our items emerge without expiration, by assuming all items with the same degree of tolerance of delay, this algorithm will select racks whose items are emerged earliest.
    \item \textbf{Integer Linear Programming planning (ILP) \cite{EJOR17Boysen}}.
    This method proposes an integer linear programming based approach to handle orders composed of items from different racks \cite{EJOR17Boysen}.
    We extend their method to our problem by adding the pickers' status in the linear programming model.
    \item \textbf{Adaptive Task Planning (ATP)}.
    This algorithm incorporates reinforcement learning for rack selection and A* algorithm for path finding as introduced in \secref{sec:atp}.
    \item \textbf{Efficient Adaptive Task Planning (\sysname)}.
    This algorithm incorporates all efficient design for both selection and planning (\algref{alg:eatp}).
\end{itemize}

\fakeparagraph{Evaluation Metrics}
We use three metrics to evaluate the algorithms in terms of \textit{effectiveness}.
\begin{itemize}
    \item \textbf{Makespan (M)}.
    It is the objective of TPRW problem as defined in \equref{eq:makespan}.
    A smaller makespan indicates a higher processing efficiency of a warehouse.
    \item \textbf{Picker's Processing Rate (PPR)}.
    This metric is defined as follows.
    \begin{equation}
    PPR = \frac{1}{|P|} \sum_{p\in P}{\frac{\sum_{t}\mathbb{I}_{p \text{ is processing at } t}}{M}}
    \end{equation}
    where $\mathbb{I}_{p \text{ is working at } t}$ is 1 if picker $p$ is processing or 0 otherwise, so the summation term in numerator is the picker's total processing time.
    A larger PPR means that all pickers are working sufficiently, leading to a higher processing efficiency.
    \item \textbf{Robot's Working Rate (RWR)}.
    Similar to PPR, RWR can be defined as follows.
    \begin{equation}
    RWR = \frac{1}{|A|} \sum_{a\in A}{\frac{\sum_{t}\mathbb{I}_{a \text{ is working at } t}}{M}}
    \end{equation}
    where the summation term in numerator is the robot's working time.
    A large RWR means that the assignment algorithm using robot in a more effective way, that is less delivering time and more picking time.
\end{itemize}

Apart from effectiveness metrics, we adopt three metrics to quantify the time and memory \textit{efficiency}.
\begin{itemize}
    \item \textbf{Selection Time Consumption (STC)}.
    The selection time consumption is the total time usage when executing the algorithms for making rack selection decisions.
    A smaller STC means a higher executing efficiency.
    \item \textbf{Planning Time Consumption (PTC)}.
    Similar to STC, the planning time consumption is the total time usage of path planning scheme generation.
    A smaller PTC means a higher executing efficiency.
    \item \textbf{Memory Consumption (MC)}.
    It measures the memory consumption when executing an algorithm.
    A smaller memory consumption indicates a higher space efficiency.
\end{itemize}

\subsection{Experimental Results}

\label{subsec:expResults}
\begin{table}[t]
	\centering
	\caption{Makespan comparison on all datasets.}
	\label{tab:makespan}
	\small
	\begin{tabular}{ccccc}
	    \toprule
		Method    & Syn-A & Syn-B & Real-Norm  & Real-Large	\\
		\midrule
		NTP \cite{AAMAS17Ma} & $95,713$ & $229,865$ & $222,044$ & $264,139$ \\
		LEF \cite{GEO16Deng} & $68,736$ & $225,484$ & $176,317$ & $-$ \\
		ILP \cite{EJOR17Boysen} & $72,423$ & $219,555$ & $173,446$ & $-$ \\
		\makecell[c]{ATP \\ (Ours)}& $\mathbf{60,193}$ & $\mathbf{209,531}$ & $\mathbf{165,438}$ & $\mathbf{220,257}$ \\
		\makecell[c]{\sysname \\ (Ours)} & $\mathbf{60,753}$ & $\mathbf{209,866}$ & $\mathbf{164,628}$ & $\mathbf{220,263}$ \\ 
		\bottomrule
	\end{tabular}
\end{table}

\fakeparagraph{Overall Performance}
The makespan results of all datasets are shown in \tabref{tab:makespan}.
ILP and LEF are too slow to execute on dataset Real-Large so we only compare our methods with NTP.
Our ATP and \sysname reduce makespan by 4.5\% \textasciitilde 37.1\% than the baselines.
Specifically, our \sysname outperforms LEF, ILP by 8.5\% and 8.7\% on average, respectively.

\figref{fig:pprrwr} illustrates the comparison results for other effectiveness metrics.
The x-axis (y-axis) indicates the task planning procedure (the value of PPR or RWR).
Our \sysname and ATP achieves the highest PPR and RWR, which is 4.6\% \textasciitilde 59.1\% higher and 3.5\% \textasciitilde 59.8\% than baselines, respectively.
In particular, our \sysname outperforms LEF by 9.4\% and 9.3\% on average in terms of PPR and RWR, respectively.
As for ILP, \sysname gains an average improvement of 9.9\% and 10.2\% in terms of PPR and RWR, respectively.
Also, \sysname consistently outperforms the baselines in all the three datasets.

We also find an interesting fact that PPR has nearly the same variations as RWR.
This is reasonable because all pickers' processing workload is delivered by robots thus robots are as busy as pickers.
Also note that \sysname has a slightly performance loss than ATP.
This is because the adoption of efficient design introduced in \secref{sec:efficient} trades some precision for acceleration, which leads to a slightly performance loss.
However, the performance loss is less than 1\% and it still significantly outperforms other baselines, so the trade-off is worthwhile.
We will show that \sysname has a large reduction on both time and space consumption.

\fakeparagraph{Adaptibility}
Our ATP and \sysname shows a strongly adaptive property mainly because:  \textit{(i)} our algorithms are steadily outperforms other baselines overall datasets that have different layouts, levels of throughput and so on, indicating our algorithm are capable of multiple situations. 
\textit{(ii)} PPR and RWR of our algorithms steadily remain high during the task planning procedure, unlike other baselines vary largely (See \figref{fig:pprrwr}), which implies that our algorithms can adaptively make task planning decisions and keep robots and pickers working steadily despite of the time-variant throughput.

\begin{figure*}[t]
\centering
	\subfigure[STC on Syn-A]{
		\includegraphics[width=0.23\linewidth]{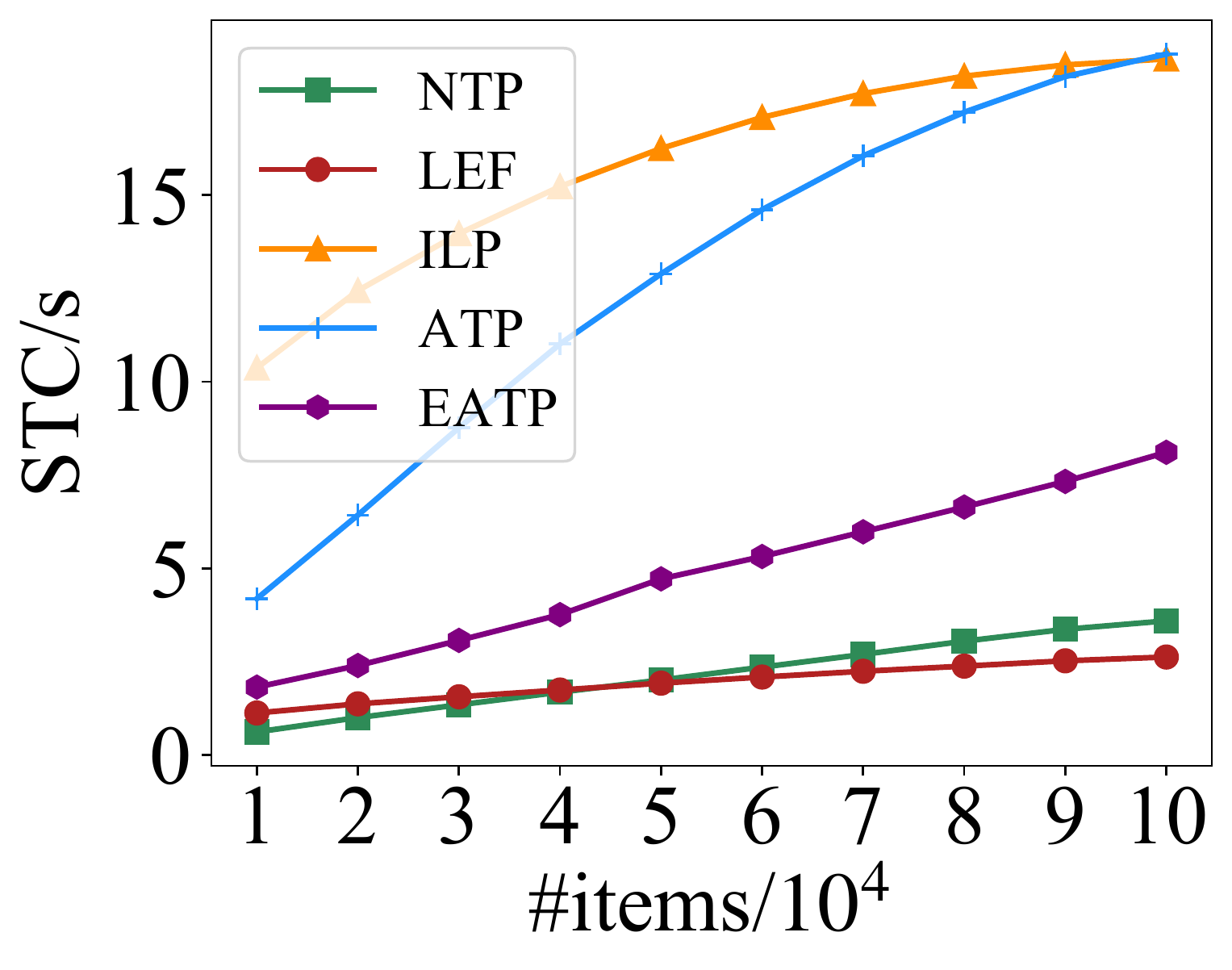}
		\label{subfig:normalSTC}
	}
	\subfigure[STC on Syn-B]{
		\includegraphics[width=0.23\linewidth]{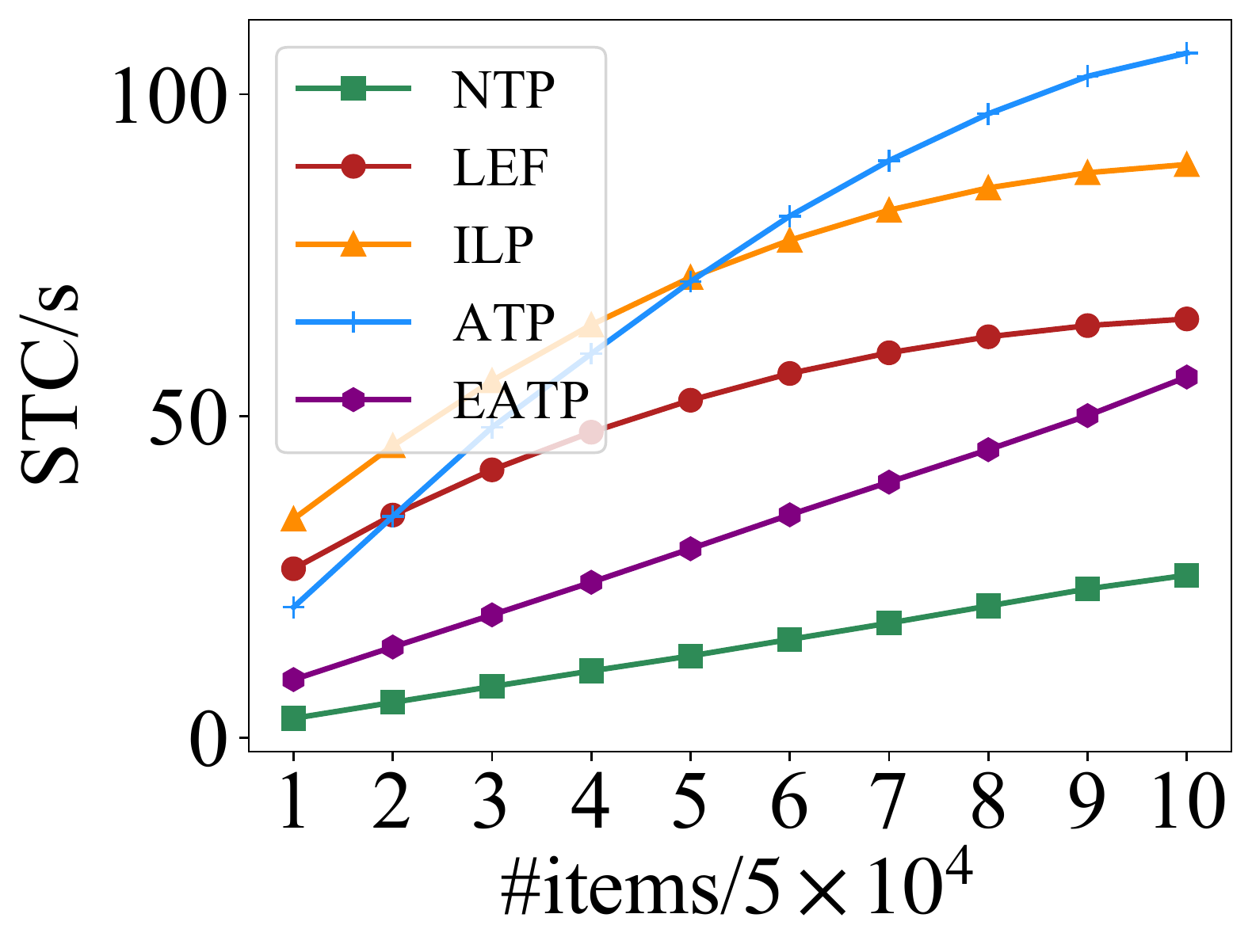}
		\label{subfig:largeSTC}
	}
	\subfigure[STC on Real-Norm]{
		\includegraphics[width=0.23\linewidth]{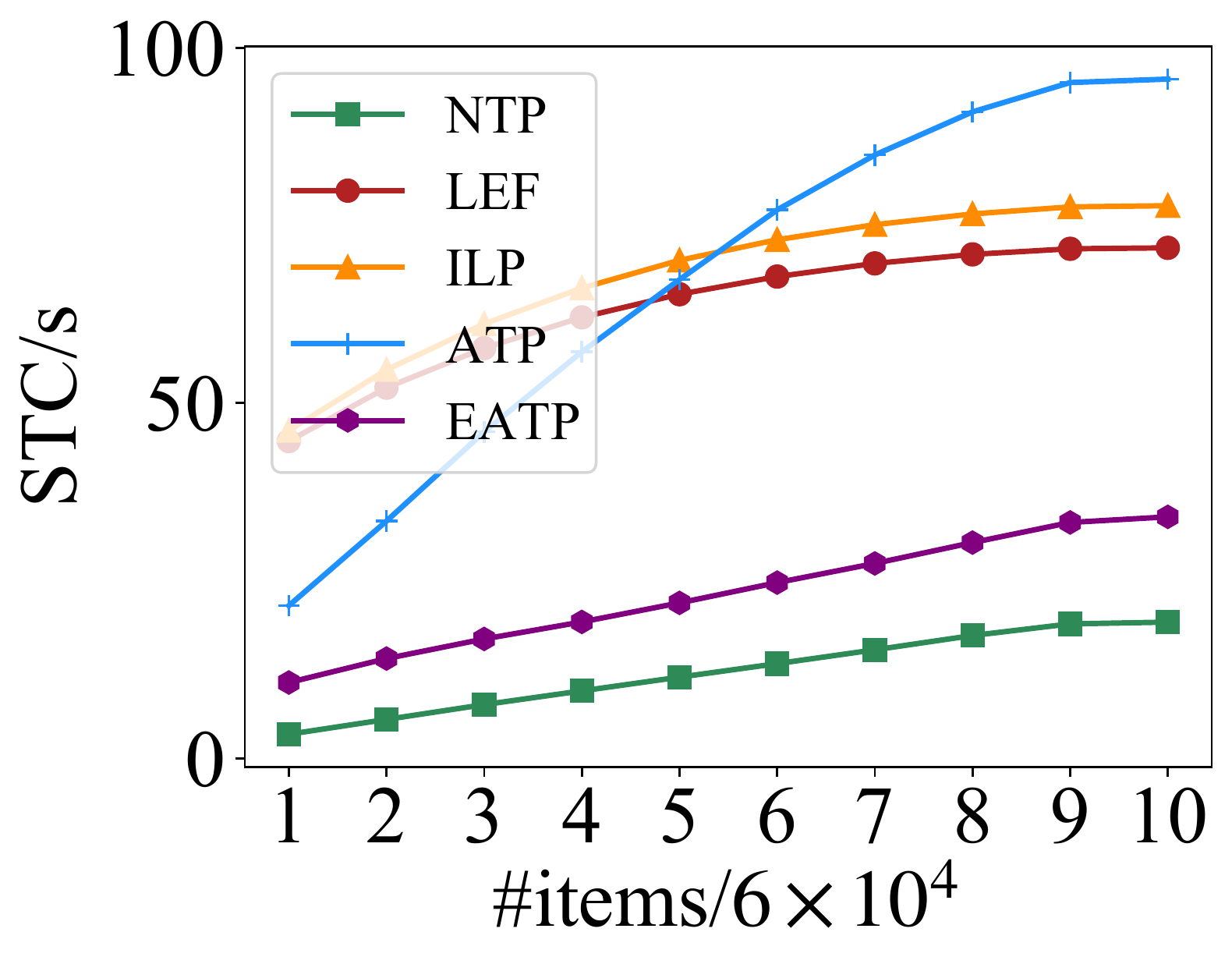}
		\label{subfig:realSTC}
	}
	\subfigure[STC on Real-Large]{
		\includegraphics[width=0.23\linewidth]{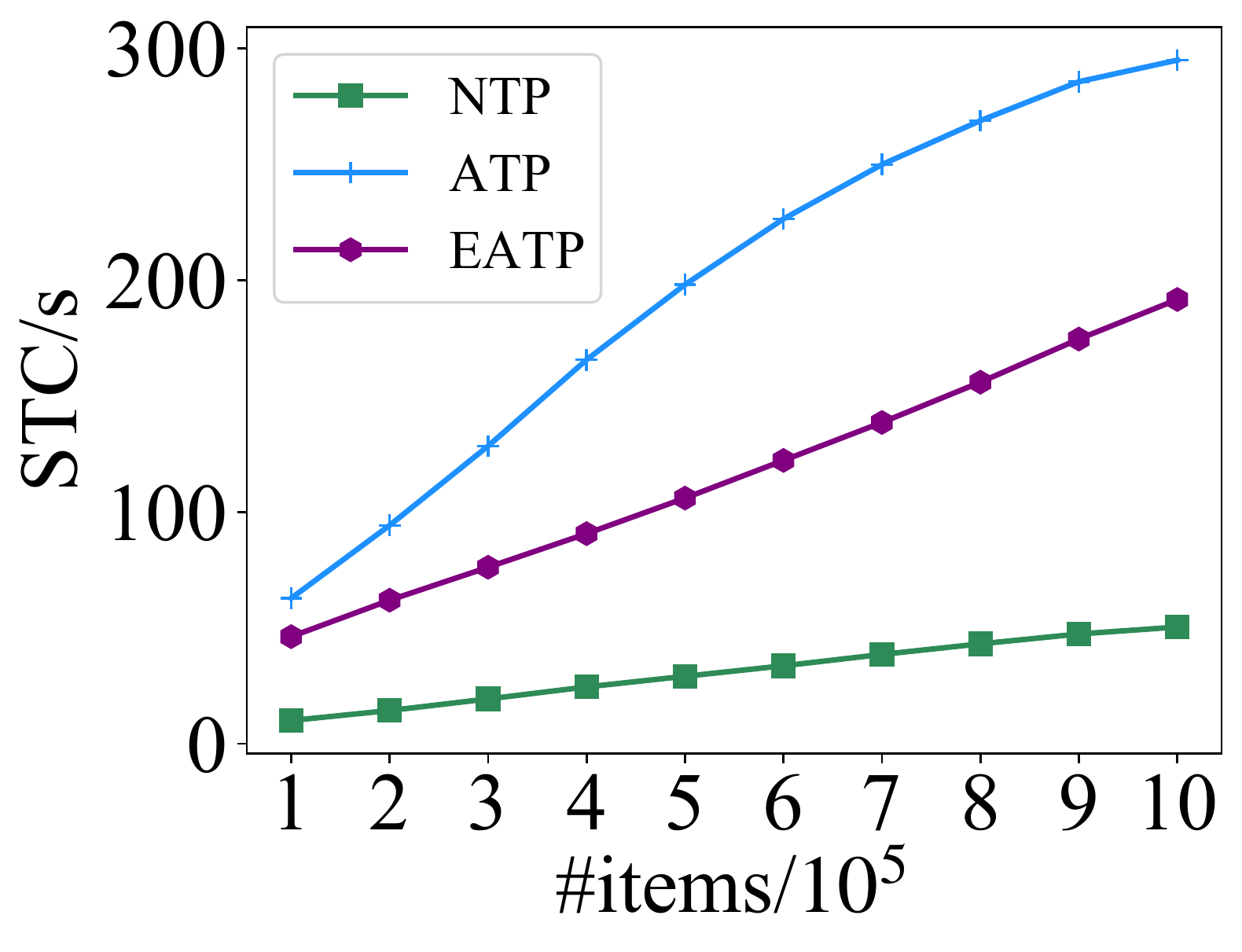}
		\label{subfig:superSTC}
	}
	\subfigure[PTC on Syn-A]{
		\includegraphics[width=0.23\linewidth]{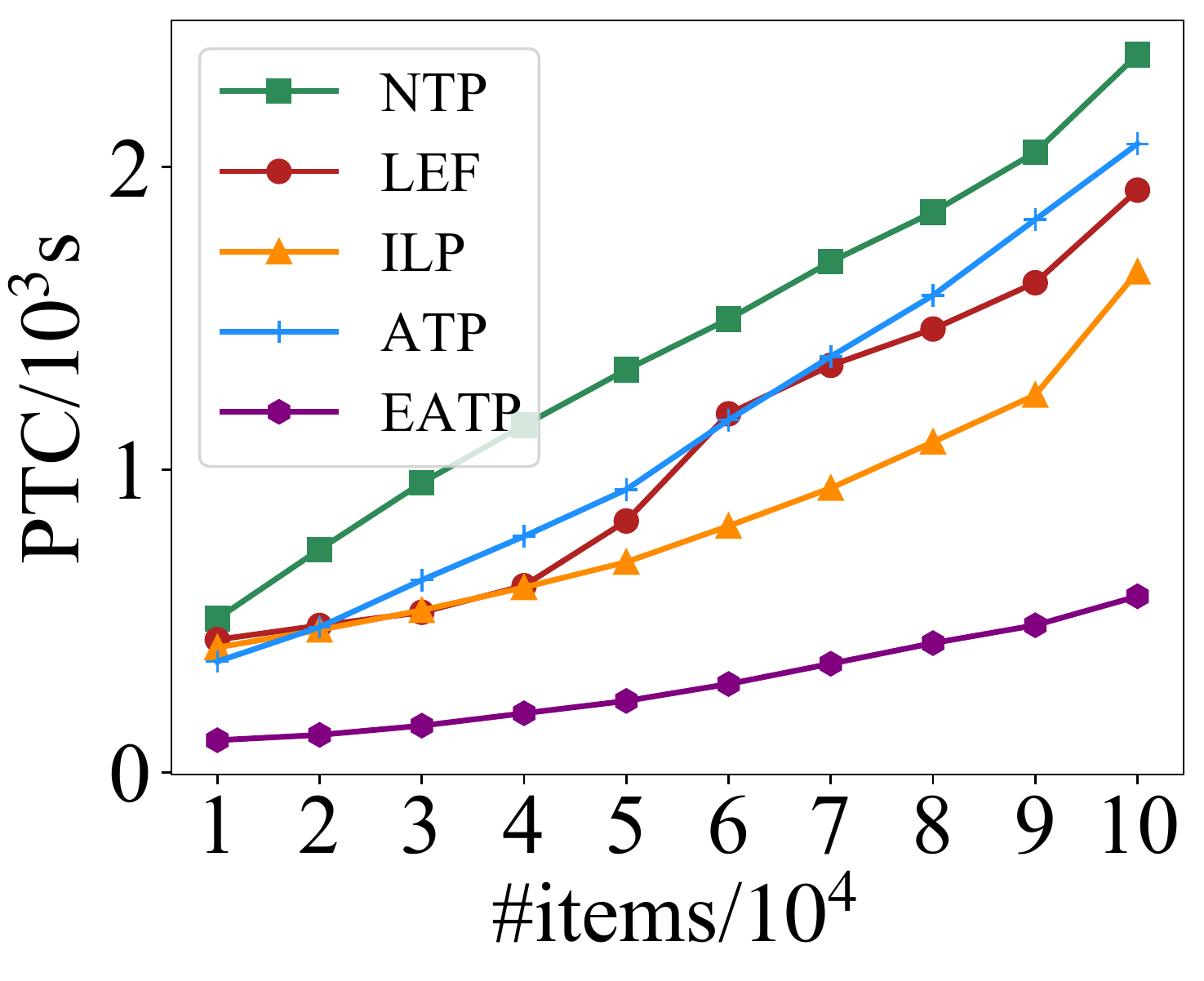}
		\label{subfig:normalPTC}
	}
	\subfigure[PTC on Syn-B]{
		\includegraphics[width=0.23\linewidth]{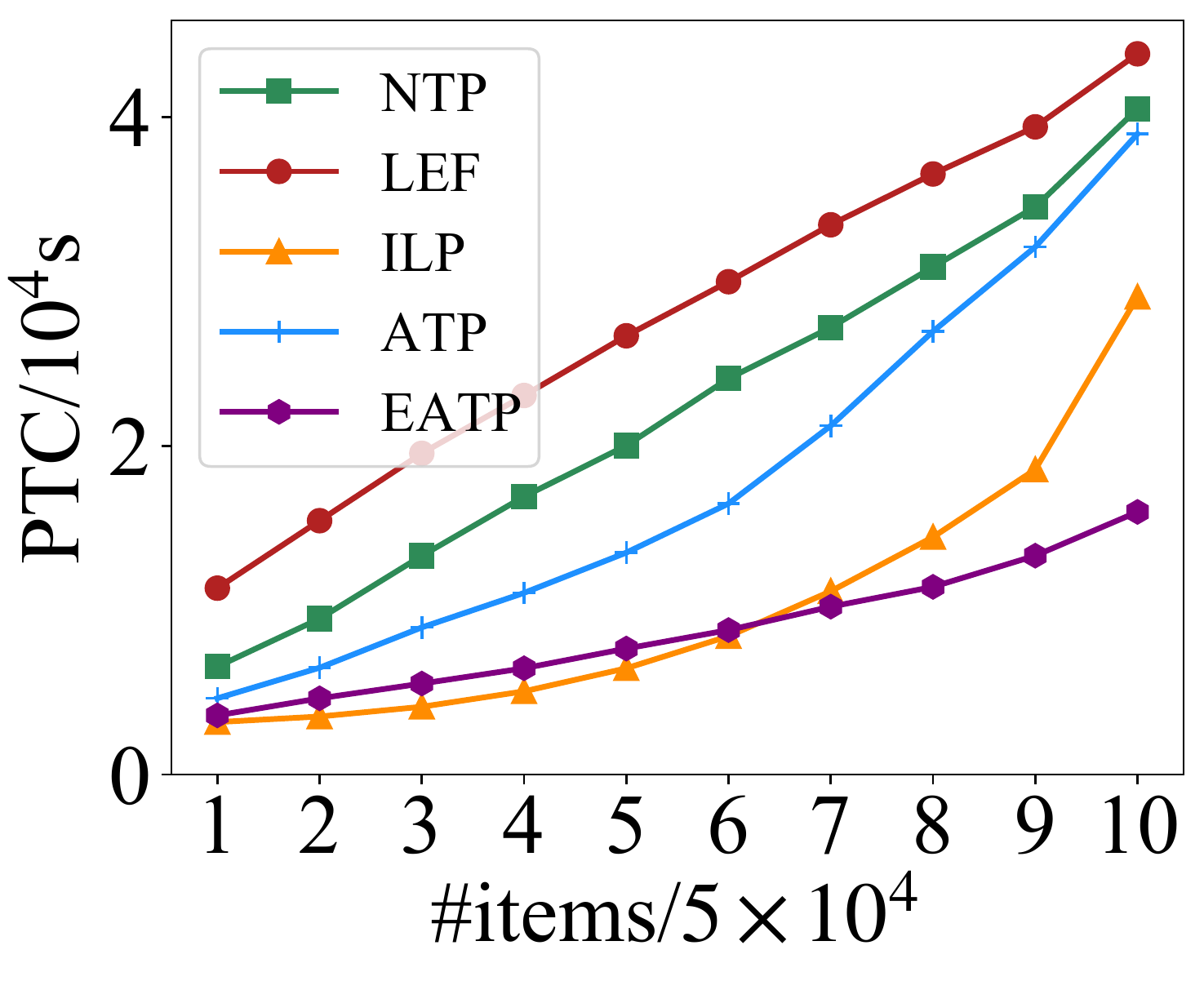}
		\label{subfig:largePTC}
	}
	\subfigure[PTC on Real-Norm]{
		\includegraphics[width=0.23\linewidth]{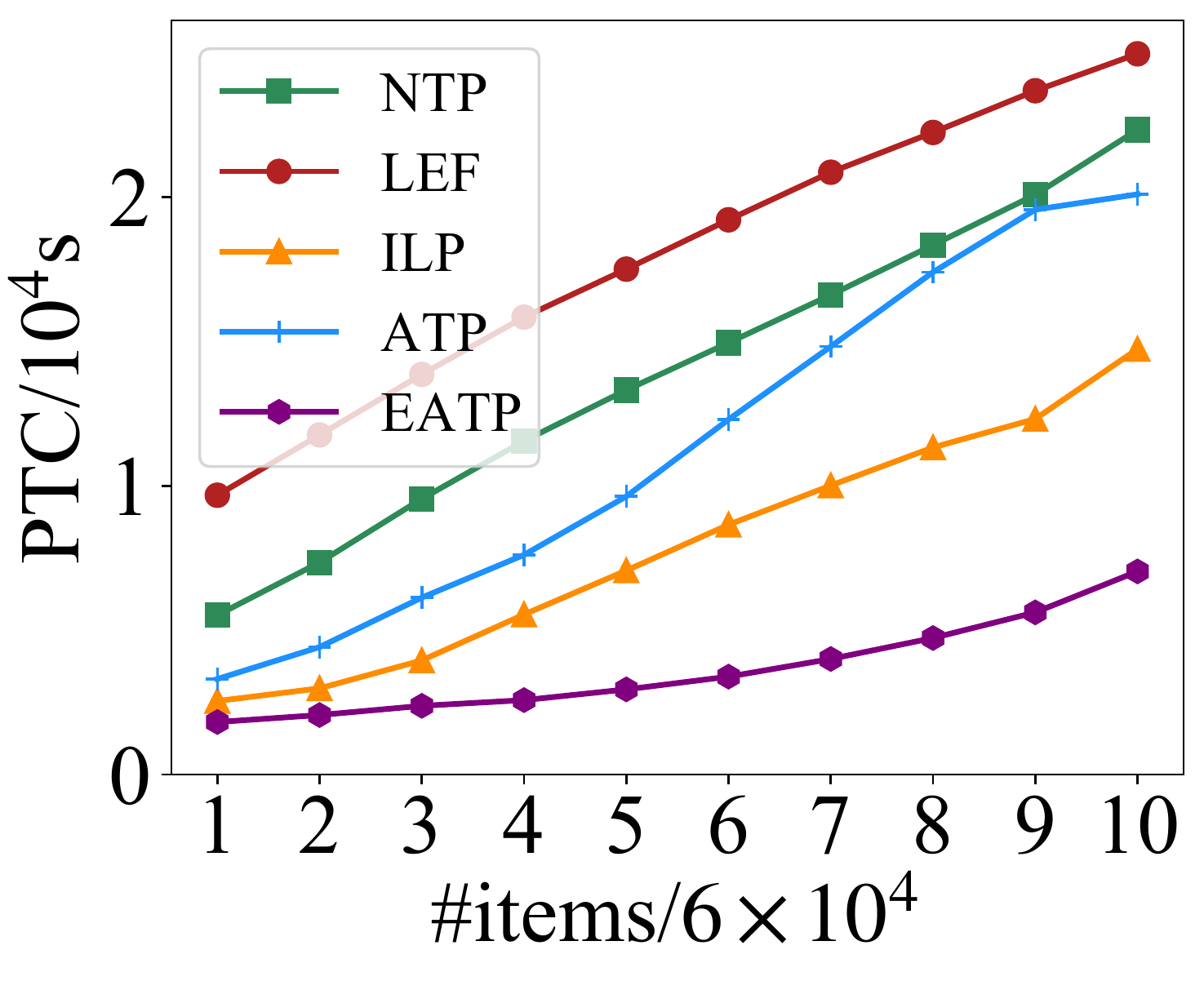}
		\label{subfig:realPTC}
	}
	\subfigure[PTC on Real-Large]{
		\includegraphics[width=0.23\linewidth]{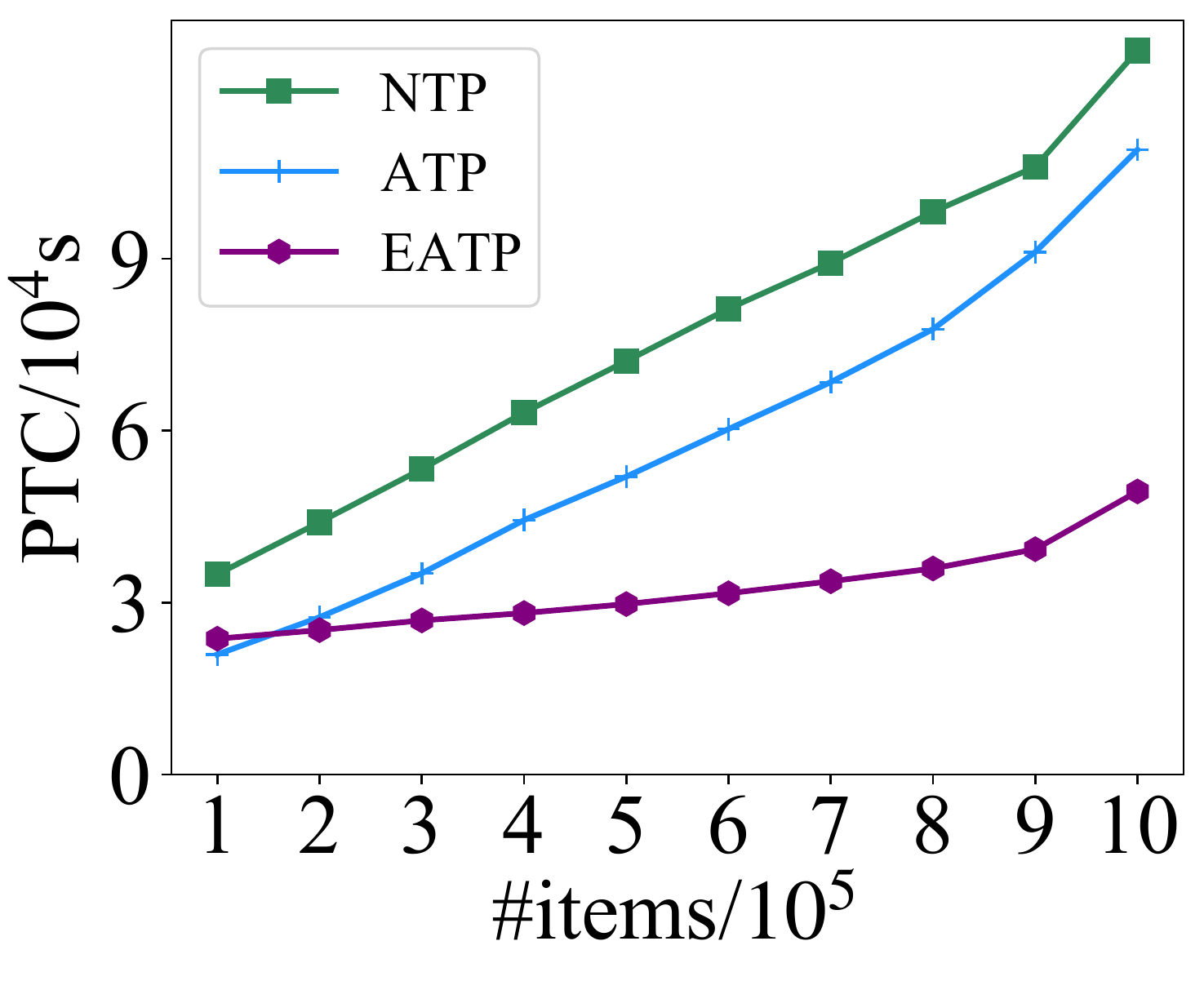}
		\label{subfig:superPTC}
	}
	\caption{Selection Time Consumption (STC) and Planning Time Consumption (PTC) comparisons.}
	\label{fig:stcptc}
\end{figure*}

\begin{figure*}[t]
\centering
	\subfigure[Syn-A]{
		\includegraphics[width=0.23\linewidth]{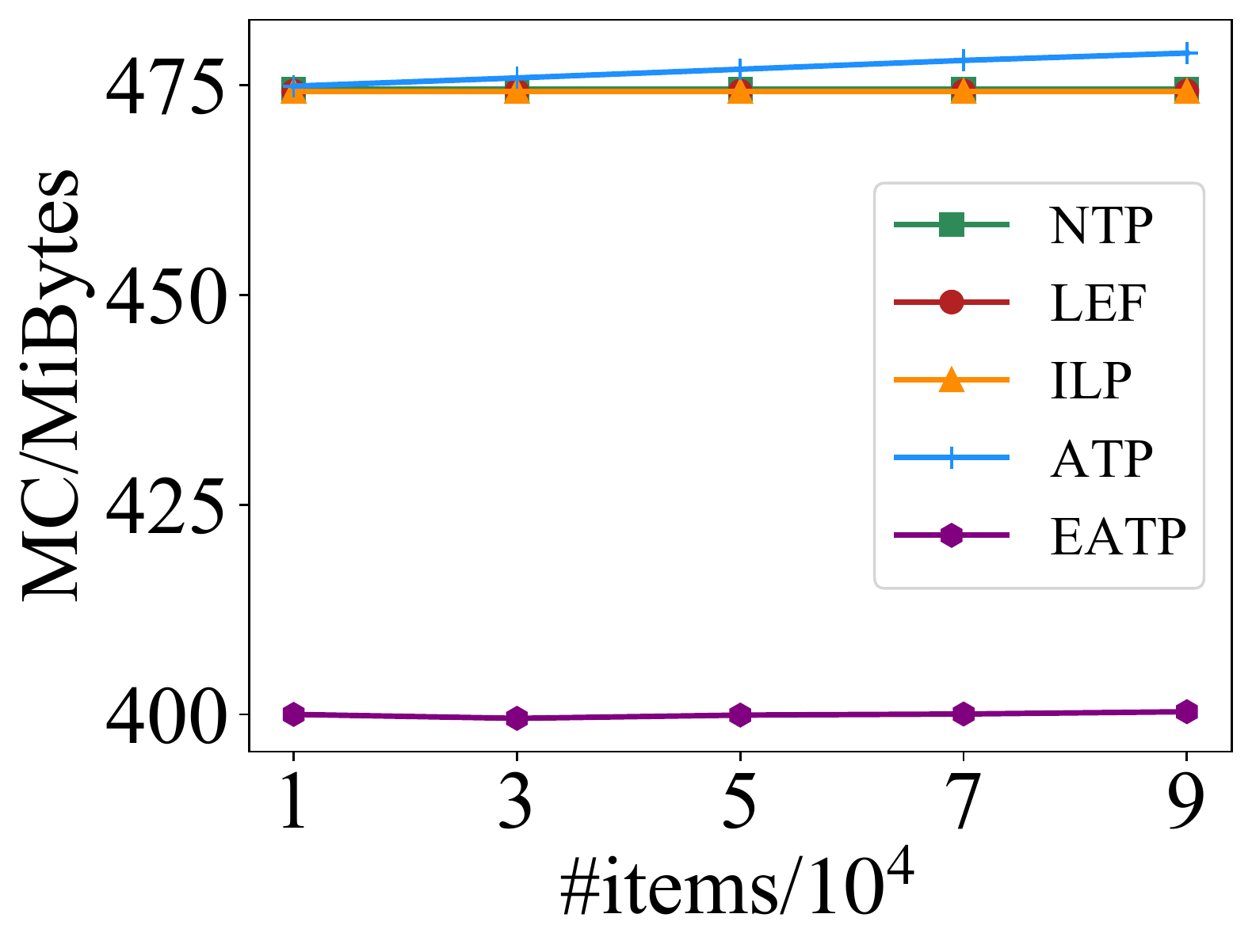}
		\label{subfig:normalMC}
	}
	\subfigure[Syn-B]{
		\includegraphics[width=0.23\linewidth]{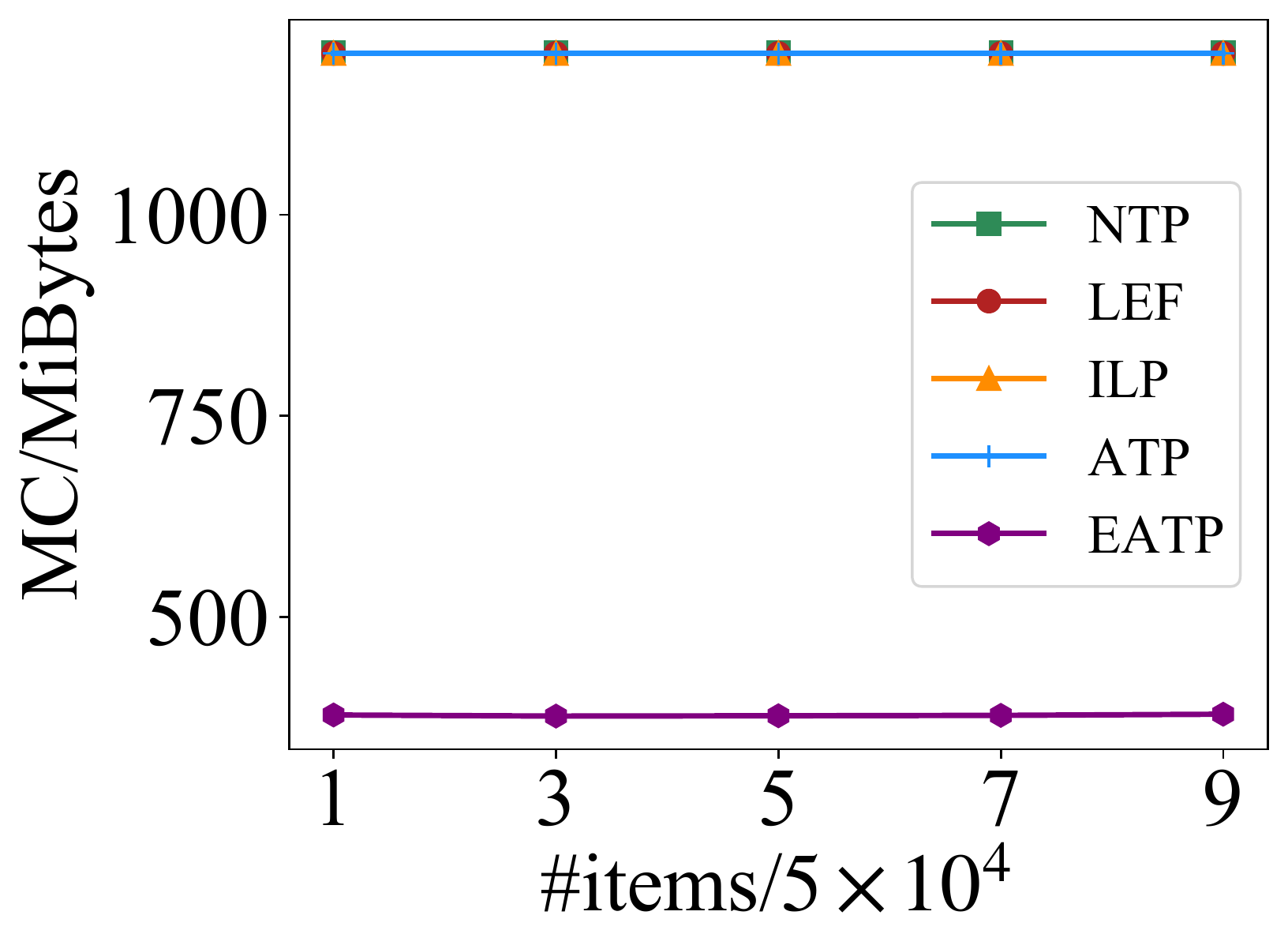}
		\label{subfig:largeMC}
	}
	\subfigure[Real-Norm]{
		\includegraphics[width=0.23\linewidth]{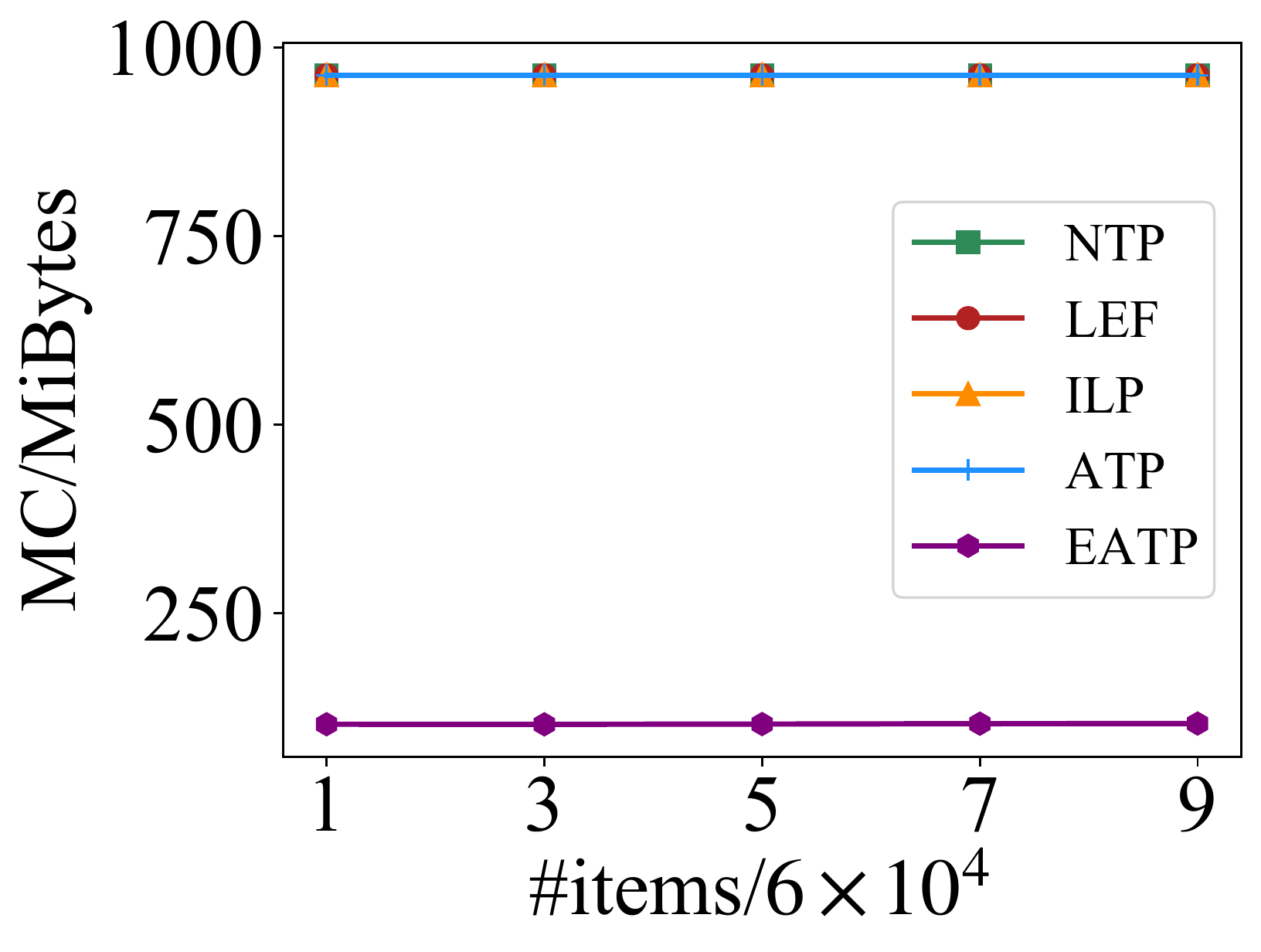}
		\label{subfig:realMC}
	}
	\subfigure[Real-Large]{
		\includegraphics[width=0.23\linewidth]{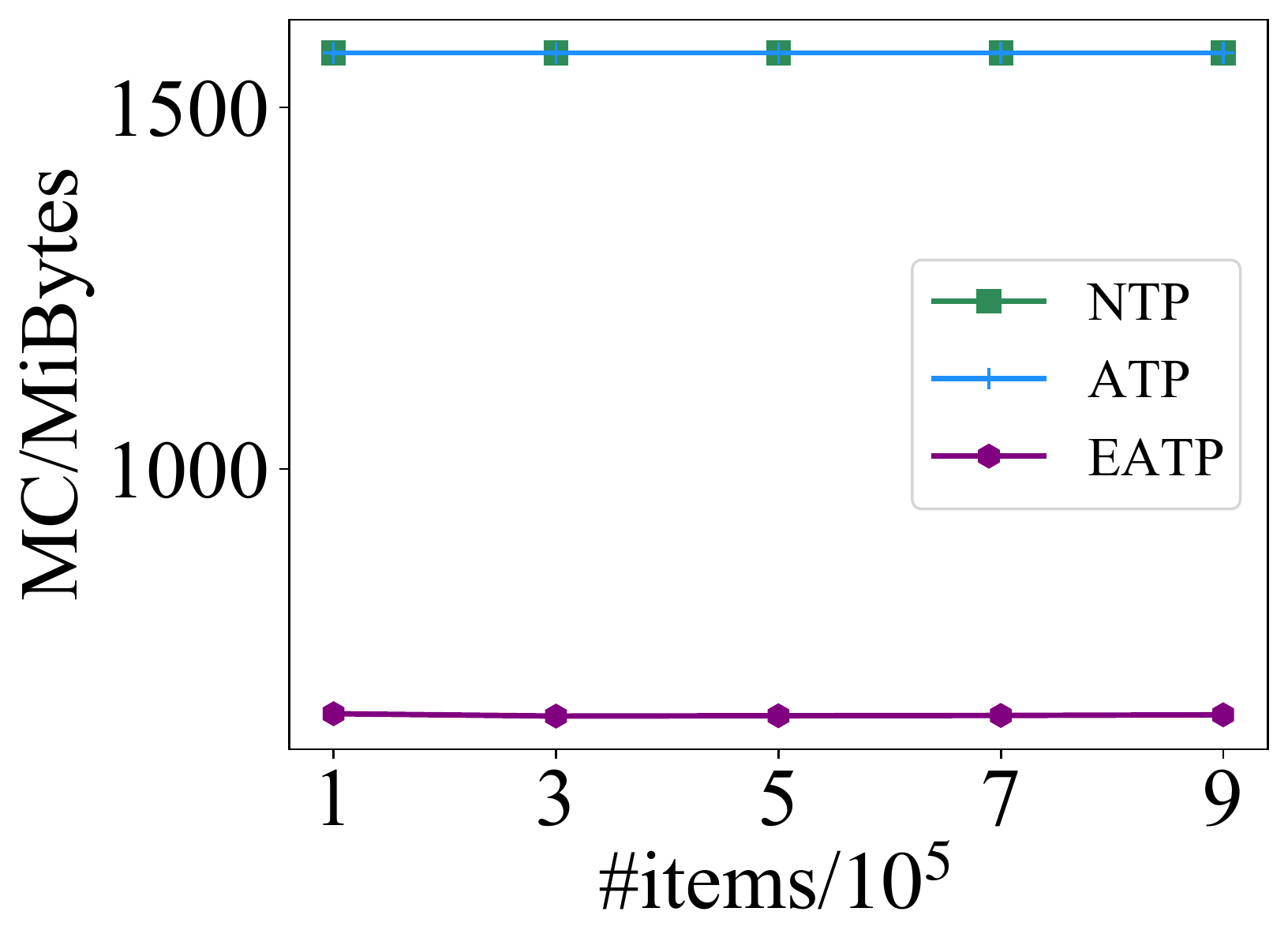}
		\label{subfig:superMC}
	}
	\caption{Memory Consumption comparisons.}
	\label{fig:mc}
\end{figure*}

\fakeparagraph{Scalability}
\figref{fig:stcptc} illustrates the selection and planning time consumption results.
For selection time consumption, without adoption of efficient design, ATP does not perform well and it is even worse than ILP.
With the selection optimization (\secref{subsec:effSelection}), our \sysname's efficiency improves significantly by 153.8\% \textasciitilde 280.9\% and close to most naive Greedy methods, whose complexity is only $O(|P|\log|P|+|A|)$.
Its time usage is less than LEF and ILP by 52.7\% and 56.5\% at most, respectively.
As for planning time consumption, our efficient design introduced in \secref{subsec:effFinding} improves \sysname planning efficiency largely.
Specifically, our \sysname has a reduction of 75.5\%, 60.5\%, 71.8\% and 60.8\% at most compared with other algorithms on dataset Syn-A, Syn-B Real-Normal and Real-Large respectively.
This results shows that our algorithm has a strong potential for large-scale processing requirement.
Especially on Real-Large, our \sysname's total execution time are less than NTP for over 7000 seconds.

Note that even though ATP adopts the same path planning algorithm as other baselines, it still has a smaller PTC because its adaptibility helps reduces the planning frequency.

As for memory cost, \figref{fig:mc} illustrates the comparison results.
All algorithms except \sysname have nearly the same memory cost due to the memory cost bottleneck lies on A*-based planning which these algorithm both adopts.
Besides, all algorithm has a steadily usage of memory as the task planning procedure goes on, this is because we eliminate passed saptiotemporal graph or timestamps timely and maintain the memory consumption relatively stable.
The comparison results shows our \sysname is also memory efficient.
With the help of conflict detection table (\secref{subsec:effFinding}), we can largely reduces the memory usage by 16.4\%, 68.4\%, 89.2\% and 58.1\% on Syn-A, Syn-B, Real-Normal and Real-Large, respectively.

% \begin{figure}[h]
%     \centering
%     \subfigure[Makespan variations over $\delta$]{
% 		\includegraphics[width=0.4\linewidth]{fig/parameter_delta.pdf}
% 		\label{subfig:parameterDelta}
% 	}
%     \subfigure[PTC variations over $L$]{
% 		\includegraphics[width=0.42\linewidth]{fig/parameter_L.pdf}
% 		\label{subfig:parameterL}
% 	}
% 	\caption{validation on hyperparameter $\delta$ and $L$}
%     \label{fig:parameter}
% \end{figure}

% \fakeparagraph{Impact of Hyperparameters}
% We studies impact of two hyperparameters $\delta$ and $L$, which control degree of bootstrap and cache-aiding, respectively.
% A smaller $\delta$ encourages more to bootstrap operations while a larger $L$ encourages using the cached path instead of finding the shortest path indicating trading for more efficiency.
% It turns out that neither too small nor too large $\delta$ will lead to a good result.
% The best value of $\delta$ lies between 0.2 and 0.4 (\figref{subfig:parameterDelta}).
% Greedy provides a rough estimation on those unexplored states, which helps the bootstrap train the value function (when $\delta$ is small).
% However, this estimation is too rough and cannot be relied on too much, otherwise the performance is bad (when $\delta$ is large). 
% \figref{subfig:parameterL} certificates that as $L$ grower, the PTC decreases by over 100s.

\fakeparagraph{Summary of Experimental Results} 
The experimental results are summarized as below.
\begin{itemize}
    \item 
    Our ATP and \sysname achieve state-of-the-art effectiveness performances.
    With an reduction on makespan by mostly 37.1\% than other baselines.
    \item 
    The steady superiority over other baselines on all datasets and the steadily high PPR and RWR during the whole task planning procedure indicates a high adaptability of our algorithm.
    \item 
    The efficiency design on both selection and planning improves efficiency by up to 280.9\%, which enables our \sysname to overcome scalability challenge.
    % \item 
    % Suitable hyperparameters $\delta$ and $L$ contribute to better effectiveness or efficiency.
\end{itemize}

% \subsection{Case Study of Bottleneck Variations}
\subsection{Case Study of Bottleneck Variations on Geekplus}
\label{subsec:case}
\begin{figure}[t]
    \centering
	\includegraphics[width=0.9\linewidth]{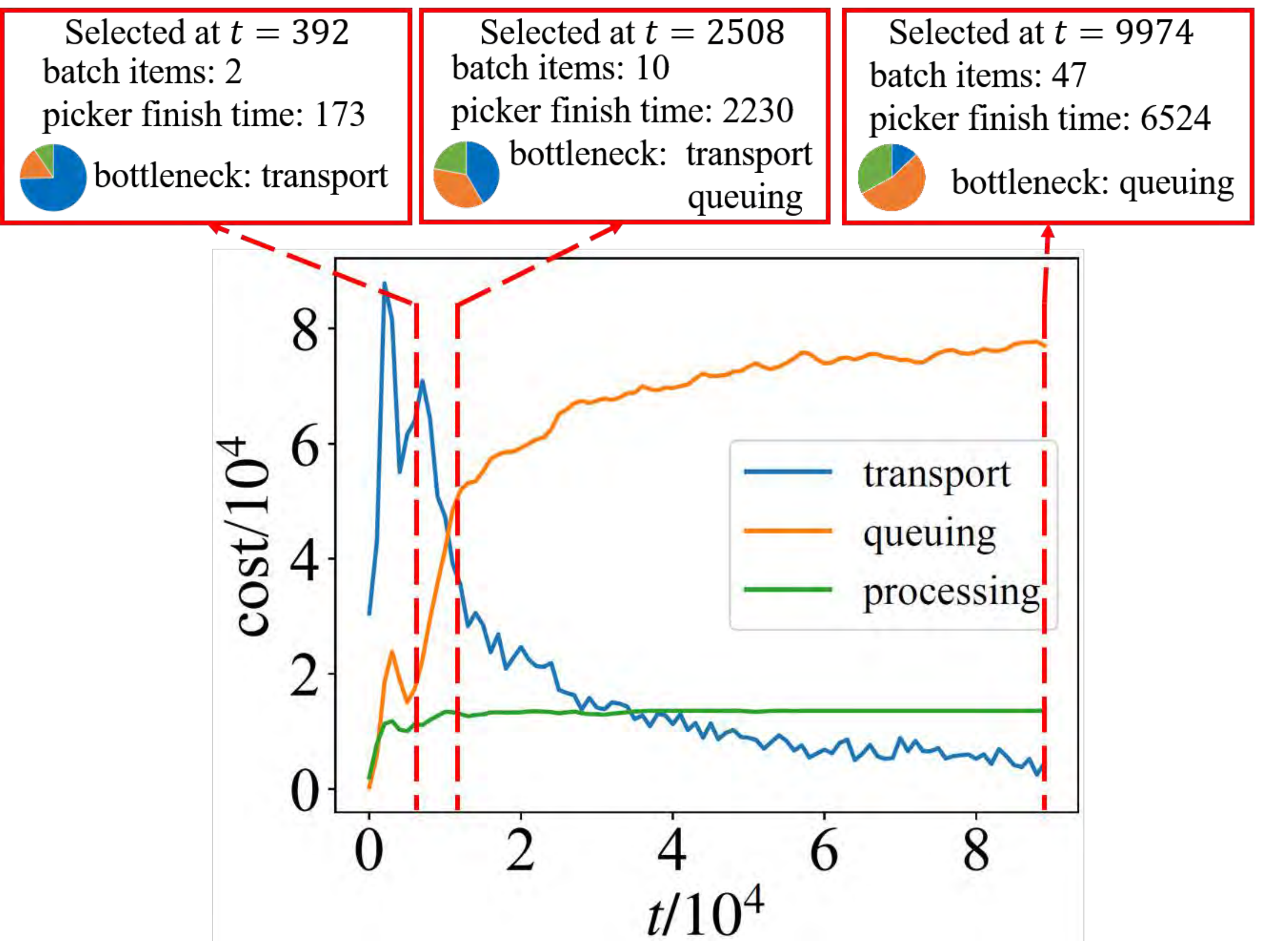}
	\caption{Bottleneck variation over time.}
    \label{fig:bottleneck}
\end{figure}

% \fakeparagraph{Bottleneck variation}
We conduct case study on demonstrative warehouse with over 1 thousand robots and 50 thousand items which is built by Geekplus\textsuperscript{\ref{geek}}, a leading smart logistics company.
We validate the phenomenon of bottleneck variations and how our method can adaptively batching items.
The bottleneck variations among processing, queuing and transport (\ie summation of pickup, delivery and return) are shown in \figref{fig:bottleneck}.
The x-axis is picking time while y-axis is cost summation of all racks under different fulfillment steps.
At the beginning, when the number of items is small, the bottleneck lies in transport.
As times goes on, the number of items is continually growing, bottleneck convert to queuing.
Meanwhile the processing time grows and then it remains static.

Next we choose a single rack to illustrate how our ATP accounts for the situation and make decisions adaptively.
Different items emerge on this rack at different time, where the bottleneck is different.
Our ATP decides to batch items instead of delivery immediately.
When the bottleneck lies in transport, ATP tends to batch less items while it will tends to batch more orders as queuing time becomes larger.

\section{Related work}
\label{sec:related}
Our study is related to two threads of studies.

\subsection{Multi-Agent Path Finding}
Multi-agent path finding is inspired by the need for coordinating hundreds of robots in a robotized warehouse \cite{AI08Wurman}.
The problem is about searching for shortest paths while meeting up with conflict constraints \cite{SOCS19Stern}.
Although this problem has been proved to be NP-hard \cite{AAAI10Surynek}, earlier studies still try to design branch-bound search algorithms hoping to find the optimal solution in an offline setting \cite{AAAI10Standley, AI13Sharon, AI15Sharon, IJCAI15Boyarski}.
Yet these solutions are unfit for large-scale applications.
For example, the empirical validations in these studies typically consider fewer than 200 robots and delivery tasks, while their running time can be up to 5 minutes, which fails to support online planning with over thousands robots and million-scale daily throughput.

More recently, online multi-agent path finding has been explored \cite{AAMAS17Ma, AAAI19vancara, AAAI21Li}.
In \cite{AAMAS17Ma}, the authors study the online multi-agent path finding problem where tasks emerge in an online manner while robots remain static, which is same as our settings.
Other efforts \cite{AAAI19vancara, AAAI21Li} consider a different setting where robots and tasks are tied and emerge together.
The setting ignores the process to assign robots to tasks, which is not aligned with the rack-to-picker setting in our problem.

In this work, we study online multi-agent path finding in settings more aligned with large-scale applications.
On the one hand, we aim to minimize the end-to-end makespan that covers the entire fulfilment cycle (pickup, delivery, queuing, processing, and return).
On the other hand, we account for the high-volume, high-variation item arrivals.
Existing solutions fail to deliver satisfactory effectiveness and efficiency in these two settings.
%Though conversion from offline to online makes the problem more align with real scenario,
%these solutions, however, still focus only on the optimization of transportation time.
%These optimization may be effective only when the makespan bottleneck is constantly the transport time (pickup, delivery and return).
%However, dynamic makespan bottleneck dwarfs performances of these solutions.

\subsection{Task Assignment and Planning in Spatial Crowdsourcing}
% task assignment
Task assignment and planning are two core issues in spatial crowdsourcing \cite{VLDB20Tong}, a popular topic in database research.

Task assignment is typically modeled as a bipartite matching problem, where vertices from two sides represented tasks and workers, respectively \cite{ICDE16Tong}.
Various optimization goals have been investigated, including maximizing the matching edge weights \cite{KDD18Xu}, minimizing the detour distance \cite{GIS20Camila} and load balancing \cite{ICDE21Zhao, VLDB20Chen}.
Alternatively, task assignment can be viewed as a selection problem where the objective is to maximize the number of fulfilled tasks \cite{GEO16Deng, GIS13Deng}.

Although these formulations are aligned with different spatial crowdsourcing applications such as ride hailing \cite{KDD18Xu, VLDB20Chen}, geographical data generation \cite{GIS20Camila, GEO16Deng, GIS13Deng} and workload distribution \cite{ICDE21Zhao, VLDB20Chen}, they cannot be trivially extended into our problem setting.
This is because spatial crowdsourcing confines the assignment decision into a batch or assume an expiration of each task.
After expiration, a task is discarded.
In our scenario, all tasks must be fulfilled and assignment decisions cover the whole time horizon.
%which adds the optimization difficulty.

% planning
Task planning, or route planning, as another important issue in spatial crowdsourcing \cite{ICDE10Hans, VLDB18Tong, ICDE13Ma, VLDB20Zeng, ICDE21Li}.
These studies optimize different objectives \cite{VLDB20Zeng, ICDE10Hans, VLDB18Tong} and also adopt data-driven approaches \cite{ICDE21Li, ICDE13Ma} to improve the planning performance.
Notice that \cite{ICDE21Li} is also related to logistics and uses reinforcement learning.
Their methods cannot be adapted to our problem because they focus on planning \textit{among} warehouses and factories on road networks without considering conflicts, and the reinforcement learning are used to avoid myopic optimization of the travel distance, rather than the bottleneck variations in our problem.
Furthermore, the number of tasks and robots we plan is ten times larger than their orders and vehicles, making us confronting stricter efficiency requirement.
Since works in spatial crowdsourcing are humans, their route planning does not consider issues such as conflicts.
In contrast, route planning for robotized warehouses is more challenging for the coordination among robots to avoid conflicts.
%However, route planning on road network is unfit for robotized warehouse mainly because of the requirement for coordination among robots to avoid conflict.

An orthogonal research on path planning focuses on indoor spaces, where the positioning data may be uncertain and noisy \cite{ICDE12Lu, ICDE13Xie, ICDE20Liu}.
We assume the location data of robots are accurate, which is common in real-world robotized warehouses.
Task planning under location errors is out of our scope.
%These studies stress more on model of topological layout of indoor space and overcome the uncertainty of indoor positioning data, but do not consider collisions either.
%This conflict avoidance issue has been studied by another research community, as introduced next.

\section{Conclusions}
\label{sec:conclusion}

In this paper, we propose the TPRW problem, an extension from online multi-agent path finding problem.
It defines an end-to-end makespan incorporating all steps of item fulfillment, which is suitable for large-scale and highly varied throughput in modern robotized warehouses.
Direct extension of state-of-the-art methods is inflexible to solve the problem.
In response, we propose the framework of efficient adaptive task planning (\sysname).
\sysname exploits reinforcement learning to adaptively decide and plan paths for robots.
It also adopts a set of acceleration techniques to optimize both time and memory efficiency.
Experimental results show that \sysname achieves 37.1\% and 75.5\% improvement in effectiveness and efficiency over the state-of-the-art online multi-agent path finding algorithms.

% \newpage
\section*{Acknowledgments}
\label{sec:ack}
We are grateful to anonymous reviewers for their constructive comments. 
This work is partially supported by 
the National Key Research and Development Program of China under Grant No. 2018AAA0101100, 
the National Science Foundation of China (NSFC) under Grant No. U21A20516, U1811463 and 62076017, 
and the State Key Laboratory of Software Development Environment Open Funding No. SKLSDE-2020ZX-07.
This research was supported by the Lee Kong Chian Fellowship awarded to Zimu Zhou by Singapore Management University.
Yongxin Tong is the corresponding author in this paper.

% 【童老师检查参考】
% 当前基金号对应：
% 2018AAA0101100 国家重点研发计划
% 61822201 优青项目
% U21A20516 新中的轨道项目
% U1811463 吕校长广东大基金
% 62076017 徐导项目，一般都挂
% 61690202 不掌握是哪个项目，但老师上次自己改的时候加上了
% Lee Kong Chian Fellowship 子慕老师基金号

% 【本次考虑删掉的基金号】
% Z191100002519012 北京市项目，21年底已结题
% CAAIXSJLJJ-2020-020-A 华为项目，21年底已结题

% 【需着重确认】
% SKLSDE-2020ZX-07  软国重的项目，2022年是否有新的项目号

\bibliographystyle{IEEEtran}
\balance
\bibliography{ref}

\begin{thebibliography}{10}
\providecommand{\url}[1]{#1}
\csname url@samestyle\endcsname
\providecommand{\newblock}{\relax}
\providecommand{\bibinfo}[2]{#2}
\providecommand{\BIBentrySTDinterwordspacing}{\spaceskip=0pt\relax}
\providecommand{\BIBentryALTinterwordstretchfactor}{4}
\providecommand{\BIBentryALTinterwordspacing}{\spaceskip=\fontdimen2\font plus
\BIBentryALTinterwordstretchfactor\fontdimen3\font minus
  \fontdimen4\font\relax}
\providecommand{\BIBforeignlanguage}[2]{{%
\expandafter\ifx\csname l@#1\endcsname\relax
\typeout{** WARNING: IEEEtran.bst: No hyphenation pattern has been}%
\typeout{** loaded for the language `#1'. Using the pattern for}%
\typeout{** the default language instead.}%
\else
\language=\csname l@#1\endcsname
\fi
#2}}
\providecommand{\BIBdecl}{\relax}
\BIBdecl

\bibitem{IR16Robert}
R.~Bogue, ``Growth in e-commerce boosts innovation in the warehouse robot
  market,'' \emph{Ind. Robot}, vol.~43, no.~6, pp. 583--587, 2016.

\bibitem{AI08Wurman}
P.~R. Wurman, R.~D'Andrea, and M.~Mountz, ``Coordinating hundreds of
  cooperative, autonomous vehicles in warehouses,'' \emph{{AI} Mag.}, vol.~29,
  no.~1, pp. 9--20, 2008.

\bibitem{AAAI10Standley}
T.~Standley, ``Finding optimal solutions to cooperative pathfinding problems,''
  in \emph{{AAAI}}, vol.~24, no.~1, 2010.

\bibitem{AI13Sharon}
G.~Sharon, R.~Stern, M.~Goldenberg, and A.~Felner, ``The increasing cost tree
  search for optimal multi-agent pathfinding,'' \emph{Artificial Intelligence},
  vol. 195, pp. 470--495, 2013.

\bibitem{AI15Sharon}
G.~Sharon, R.~Stern, A.~Felner, and N.~R. Sturtevant, ``Conflict-based search
  for optimal multi-agent pathfinding,'' \emph{Artificial Intelligence}, vol.
  219, pp. 40--66, 2015.

\bibitem{IJCAI15Boyarski}
E.~Boyarski, A.~Felner, R.~Stern, G.~Sharon, D.~Tolpin, O.~Betzalel, and
  E.~Shimony, ``Icbs: Improved conflict-based search algorithm for multi-agent
  pathfinding,'' in \emph{{IJCAI}}, 2015.

\bibitem{AAMAS17Ma}
H.~Ma, J.~Li, T.~K.~S. Kumar, and S.~Koenig, ``Lifelong multi-agent path
  finding for online pickup and delivery tasks,'' in \emph{{AAMAS}}.\hskip 1em
  plus 0.5em minus 0.4em\relax {ACM}, 2017, pp. 837--845.

\bibitem{AAAI19vancara}
J.~{\v{S}}vancara, M.~Vlk, R.~Stern, D.~Atzmon, and R.~Bart{\'a}k, ``Online
  multi-agent pathfinding,'' in \emph{{AAAI}}, vol.~33, no.~01, 2019, pp.
  7732--7739.

\bibitem{AAAI21Li}
J.~Li, A.~Tinka, S.~Kiesel, J.~W. Durham, T.~K.~S. Kumar, and S.~Koenig,
  ``Lifelong multi-agent path finding in large-scale warehouses,'' in
  \emph{{AAAI}}.\hskip 1em plus 0.5em minus 0.4em\relax {AAAI}, 2021, pp.
  11\,272--11\,281.

\bibitem{SOCS19Stern}
R.~Stern, N.~R. Sturtevant, A.~Felner, and et~al, ``Multi-agent pathfinding:
  Definitions, variants, and benchmarks,'' in \emph{{SOCS}}.\hskip 1em plus
  0.5em minus 0.4em\relax {AAAI}, 2019, pp. 151--159.

\bibitem{TSSC68Peter}
P.~E. Hart, N.~J. Nilsson, and B.~Raphael, ``A formal basis for the heuristic
  determination of minimum cost paths,'' \emph{{IEEE} Trans. Syst. Sci.
  Cybern.}, vol.~4, no.~2, pp. 100--107, 1968.

\bibitem{EJOR17Boysen}
N.~Boysen, D.~Briskorn, and S.~Emde, ``Parts-to-picker based order processing
  in a rack-moving mobile robots environment,'' \emph{European Journal of
  Operational Research}, vol. 262, no.~2, pp. 550--562, 2017.

\bibitem{sutton2018reinforcement}
R.~S. Sutton and A.~G. Barto, \emph{Reinforcement learning: An
  introduction}.\hskip 1em plus 0.5em minus 0.4em\relax Cambridge: MIT press,
  2018.

\bibitem{watkins1989}
C.~Watkins, ``Learning from delayed rewards,'' \emph{PhD thesis, Cambridge
  University}, 1989.

\bibitem{AAMAS16Ma}
H.~Ma and S.~Koenig, ``Optimal target assignment and path finding for teams of
  agents,'' in \emph{{AAMAS}}.\hskip 1em plus 0.5em minus 0.4em\relax {ACM},
  2016, pp. 1144--1152.

\bibitem{ICAPS19Florian}
F.~Grenouilleau, W.~van Hoeve, and J.~N. Hooker, ``A multi-label a* algorithm
  for multi-agent pathfinding,'' in \emph{{ICAPS}}.\hskip 1em plus 0.5em minus
  0.4em\relax {AAAI}, 2019, pp. 181--185.

\bibitem{GEO16Deng}
D.~Deng, C.~Shahabi, U.~Demiryurek, and L.~Zhu, ``Task selection in spatial
  crowdsourcing from worker's perspective,'' \emph{GeoInformatica}, vol.~20,
  no.~3, pp. 529--568, 2016.

\bibitem{AAAI10Surynek}
P.~Surynek, ``An optimization variant of multi-robot path planning is
  intractable,'' in \emph{{AAAI}}.\hskip 1em plus 0.5em minus 0.4em\relax
  {AAAI}, 2010.

\bibitem{VLDB20Tong}
Y.~Tong, Z.~Zhou, Y.~Zeng, L.~Chen, and C.~Shahabi, ``Spatial crowdsourcing: a
  survey,'' \emph{{VLDBJ}}, vol.~29, no.~1, pp. 217--250, 2020.

\bibitem{ICDE16Tong}
Y.~Tong, J.~She, B.~Ding, L.~Wang, and L.~Chen, ``Online mobile micro-task
  allocation in spatial crowdsourcing,'' in \emph{{ICDE}}.\hskip 1em plus 0.5em
  minus 0.4em\relax {IEEE}, 2016, pp. 49--60.

\bibitem{KDD18Xu}
Z.~Xu, Z.~Li, Q.~Guan, D.~Zhang, Q.~Li, J.~Nan, and et~al, ``Large-scale order
  dispatch in on-demand ride-hailing platforms: A learning and planning
  approach,'' in \emph{{KDD}}.\hskip 1em plus 0.5em minus 0.4em\relax {ACM},
  2018, pp. 905--913.

\bibitem{GIS20Camila}
C.~F. Costa and M.~A. Nascimento, ``Online in-route task selection in spatial
  crowdsourcing,'' in \emph{{SIGSPATIAL}}.\hskip 1em plus 0.5em minus
  0.4em\relax {ACM}, 2020, pp. 239--250.

\bibitem{ICDE21Zhao}
Y.~Zhao, K.~Zheng, J.~Guo, B.~Yang, T.~B. Pedersen, and C.~S. Jensen,
  ``Fairness-aware task assignment in spatial crowdsourcing: Game-theoretic
  approaches,'' in \emph{{ICDE}}.\hskip 1em plus 0.5em minus 0.4em\relax
  {IEEE}, 2021, pp. 265--276.

\bibitem{VLDB20Chen}
Z.~Chen, P.~Cheng, L.~Chen, X.~Lin, and C.~Shahabi, ``Fair task assignment in
  spatial crowdsourcing,'' \emph{{PVLDB}}, vol.~13, no.~11, pp. 2479--2492,
  2020.

\bibitem{GIS13Deng}
D.~Deng, C.~Shahabi, and U.~Demiryurek, ``Maximizing the number of worker's
  self-selected tasks in spatial crowdsourcing,'' in \emph{{SIGSPATIAL}}.\hskip
  1em plus 0.5em minus 0.4em\relax {ACM}, 2013, pp. 314--323.

\bibitem{ICDE10Hans}
H.~Kriegel, M.~Renz, and M.~Schubert, ``Route skyline queries: {A}
  multi-preference path planning approach,'' in \emph{{ICDE}}.\hskip 1em plus
  0.5em minus 0.4em\relax {IEEE}, 2010, pp. 261--272.

\bibitem{VLDB18Tong}
Y.~Tong, Y.~Zeng, Z.~Zhou, L.~Chen, J.~Ye, and K.~Xu, ``A unified approach to
  route planning for shared mobility,'' \emph{{PVLDB}}, vol.~11, no.~11, pp.
  1633--1646, 2018.

\bibitem{ICDE13Ma}
S.~Ma, Y.~Zheng, and O.~Wolfson, ``T-share: {A} large-scale dynamic taxi
  ridesharing service,'' in \emph{{ICDE}}.\hskip 1em plus 0.5em minus
  0.4em\relax {IEEE}, 2013, pp. 410--421.

\bibitem{VLDB20Zeng}
Y.~Zeng, Y.~Tong, Y.~Song, and L.~Chen, ``The simpler the better: An indexing
  approach for shared-route planning queries,'' \emph{{PVLDB}}, vol.~13,
  no.~13, pp. 3517--3530, 2020.

\bibitem{ICDE21Li}
X.~Li, W.~Luo, M.~Yuan, J.~Wang, J.~Lu, J.~Wang, J.~L{\"{u}}, and J.~Zeng,
  ``Learning to optimize industry-scale dynamic pickup and delivery problems,''
  in \emph{{ICDE}}.\hskip 1em plus 0.5em minus 0.4em\relax {IEEE}, 2021, pp.
  2511--2522.

\bibitem{ICDE12Lu}
H.~Lu, X.~Cao, and C.~S. Jensen, ``A foundation for efficient indoor
  distance-aware query processing,'' in \emph{{ICDE}}.\hskip 1em plus 0.5em
  minus 0.4em\relax {IEEE}, 2012, pp. 438--449.

\bibitem{ICDE13Xie}
X.~Xie, H.~Lu, and T.~B. Pedersen, ``Efficient distance-aware query evaluation
  on indoor moving objects,'' in \emph{{ICDE}}.\hskip 1em plus 0.5em minus
  0.4em\relax {IEEE}, 2013, pp. 434--445.

\bibitem{ICDE20Liu}
T.~Liu, Z.~Feng, H.~Li, H.~Lu, M.~A. Cheema, H.~Cheng, and J.~Xu, ``Shortest
  path queries for indoor venues with temporal variations,'' in
  \emph{{ICDE}}.\hskip 1em plus 0.5em minus 0.4em\relax {IEEE}, 2020, pp.
  2014--2017.

\end{thebibliography}

\end{document}